%% file: manuscript_rlfm_bias_var.tex
\pdfoutput=1
\documentclass[aps,twocolumn,superscriptaddress, notitlepage,longbibliography]{revtex4-1}

\usepackage{amsmath} 
\usepackage{amsthm} 
\usepackage{amssymb}	
\usepackage{physics}
\usepackage[pdftex]{graphicx}
\usepackage{hyperref} 
\hypersetup{
    colorlinks,
    citecolor=blue,
    filecolor=black,
    linkcolor=red,
    urlcolor=blue
}
\usepackage{xcolor}
\usepackage{soul}
\usepackage{url}

\newcommand{\E}[1][]{\mathrm{E}_{#1}}
\newcommand{\Var}[1][]{\mathrm{Var}_{#1}}
\newcommand{\Cov}[1][]{\mathrm{Cov}_{#1}}
\newcommand{\Bias} {\mathrm{Bias}}
\newcommand{\test}{\mathcal{E}_{\mathrm{test}}}
\newcommand{\train}{\mathcal{E}_{\mathrm{train}}}

\newcommand{\etest}{\expval{\mathcal{E}_{\mathrm{test}}}}
\newcommand{\etrain}{\expval{\mathcal{E}_{\mathrm{train}}}}

\newcommand{\vbx}{\vec{\mathbf{x}}}
\newcommand{\vby}{\vec{\mathbf{y}}}
\newcommand{\vbz}{\vec{\mathbf{z}}}

\newcommand{\vbbeta}{\vec{\boldsymbol{\beta}}}
\newcommand{\vbeps}{\vec{\boldsymbol{\varepsilon}}}
\newcommand{\vbeta}{\vec{\boldsymbol{\eta}}}
\newcommand{\hbbeta}{\hat{\boldsymbol{\beta}}}
\newcommand{\vbxi}{\vec{\boldsymbol{\xi}}}
\newcommand{\hbw}{\hat{\mathbf{w}}}
\newcommand{\hby}{\hat{\mathbf{y}}}

\newcommand{\hba}{\hat{\mathbf{a}}}

\newcommand{\qqc}{,\qquad}

\begin{document}

\title{Bias-variance decomposition of overparameterized regression with random linear features}

\author{Jason W. Rocks}
\affiliation{Department of Physics, Boston University, Boston, Massachusetts 02215, USA}

\author{Pankaj Mehta}
\affiliation{Department of Physics, Boston University, Boston, Massachusetts 02215, USA}
\affiliation{Faculty of Computing and Data Sciences, Boston University, Boston, Massachusetts 02215, USA}

\begin{abstract}

In classical statistics, the bias-variance trade-off describes how varying a model's complexity (e.g., number of fit parameters) affects its ability to make accurate predictions.
According to this trade-off, optimal performance is achieved when a model is expressive enough to capture trends in the data, yet not so complex that it overfits idiosyncratic features of the training data.
Recently, it has become clear that this classic understanding of the bias-variance must be fundamentally revisited in light of the incredible predictive performance of ``overparameterized models'' -- models that avoid overfitting even when the number of fit parameters is large enough to perfectly fit the training data. 
Here, we present results for one of the simplest examples of an overparameterized model: regression with random linear features (i.e. a two-layer neural network with a \emph{linear} activation function).
Using the zero-temperature cavity method, we derive analytic expressions for the training error, test error, bias, and variance. We show that the linear random features model exhibits three phase transitions: two different transitions to an interpolation regime where the training error is zero, along with an additional transition between regimes with large bias and minimal bias.
Using random matrix theory, we show how each transition arises due to small nonzero eigenvalues in the Hessian matrix. Finally, we compare and contrast the phase diagram of the random \emph{linear} features model to the random \emph{nonlinear} features model and ordinary regression, highlighting the new phase transitions that result from the use of linear basis functions.
\end{abstract}

\maketitle

\section{Introduction}

One of the core concepts in modern statistics and supervised learning is the bias-variance decomposition. 
It states that the test error, the predictive performance of the model on new data, can be decomposed into three parts: bias, variance, and noise~\cite{Geman1992, Bishop2006, Mehta2019}.
The bias captures errors due to underfitting, resulting from the inability of a statistical model to sufficiently express  statistical relationships present in the data distribution.  
The variance, on the other hand, characterizes errors that result from  ``over-fitting''  unrepresentative aspects of the training data set that do not generalize (e.g., label noise). 
Finally, the noise describes irreducible errors in a test data set due to randomness in the data generating process.

In classical statistics, the bias-variance trade-off suggests that optimal predictive performance is achieved by utilizing statistical models with intermediate model complexities, balancing errors due to bias and variance.
While increasing a model's complexity (e.g., increasing the number of fit parameters) reduces bias,
it comes at the price of increasing variance.  
One of the most interesting and surprising empirical results to emerge from deep learning over the last five years is the realization that this basic intuition is fundamentally incomplete; it does not apply to ``overparameterized'' models where the number of fit parameters is large enough to perfectly fit the training data (i.e. achieve zero error on the training data set)~\cite{Zhang2017}. 

While the classic bias-variance trade-off still holds in the underparameterized regime (i.e., for models that have too few fit parameters to achieve zero training error), once a model's complexity is increased passed the interpolation threshold -- the point at which the training error goes to zero -- the test error once again decreases.
The resulting combination of a ``U-shaped'' test error in the underparameterized regime and the subsequent decrease in test error in the overparameterized regime is now commonly referred to as a ``double-descent'' curve~\cite{Belkin2019, Loog2020}. This double-descent behavior seems to be a generic property of all overparameterized supervised learning models and for this reason, has become a major area of research.

An important open question in the field is to understand the double-descent phenomena in terms of classical ideas of bias and variance. 
One fruitful approach has been to analyze analytically tractable models that exhibit the double-descent phenomena~\cite{Adlam2020, Advani2020, Ba2020, Barbier2019, Bartlett2020, Belkin2020, Bibas2019, Deng2019, DAscoli2020, DAscoli2020a, Derezinski2020, Dhifallah2020, Gerace2020, Hastie2019, Jacot2020, Kini2020, Lampinen2019, Li2020, Liang2020, Liang2020b, Liao2020, Lin2020, Mitra2019, Mei2019, Muthukumar2019, Nakkiran2019, Xu2019, Yang2020, Rocks2022}. Among, the most popular of these models are linear regression (ridge regression without basis functions) 
and the random nonlinear features model (a two-layer neural network with an arbitrary nonlinear activation function where the top layer is trained and parameters for the intermediate layer are chosen to be random but fixed)~\cite{Rocks2022}.
Here, we build upon this previous work by examining a random features model for the special case of a linear activation function (i.e., the random linear features model).
Using the zero temperature cavity method, we derive analytic expressions for the bias-variance decomposition and relate these results to the eigenvalue spectrum of the Hessian matrix.

The random linear features has been treated analytically~\cite{Advani2020, Ba2020, DAscoli2020, Lin2020, Yang2020},
with a subset of these studies attempting to carrying out bias-variance decompositions~\cite{Ba2020, Lin2020, Yang2020}.
However, these studies use non-standard definitions of bias and variance that deviate from the traditional textbook definitions~\cite{Geman1992, Bishop2006}. This choice of definition can lead to qualitatively different and difficult to reconcile results. For example, the authors of Ref.~\onlinecite{Ba2020} find that the bias diverges at the interpolation threshold, while the authors of Ref.~\onlinecite{Lin2020} find no such divergence (see Ref.~\onlinecite{Rocks2022} for an in-depth discussion). 
For this reason, here we utilize the standard definitions and carry out the bias-variance decomposition in a manner consistent with traditional definitions of these quantities in the underparameterized regime,
allowing us to identify which properties stem from the model architecture versus random sampling of the data. 
In addition, we use the zero-temperature cavity method to provide an alternative derivation of the spectrum of the Hessian matrix of the random linear features model (i.e., the spectrum of a Wishart product matrix calculated previously in Ref.~\onlinecite{Dupic2014}), allowing us to directly relate the eigenvalues of the Hessian to the double-descent phenomenon.

\subsection*{Summary of Major Results}

We briefly summarize our major results:
\begin{itemize}
\item We derive analytic expressions for the test (generalization) error, training error, bias, and variance for the random linear features model with a nonlinear data distribution using the zero-temperature cavity method.
\item We find that the behavior of this model is characterized by three distinct regimes: (i) an underparameterized regime with finite training error and large bias, (ii) a second underparameterized regime with minimal, constant bias, and (iii) an overparameterized, or interpolation, regime with zero training error. 
\item We find that the three regimes are separated by three phase transitions with two transitions to the interpolation regime, each characterized by a divergence in the test error,
and one transition between the large bias and minimal bias underparameterized regimes. Importantly, we find that the variance, but not the bias, diverges at the phase transition to the interpolation regime.
\item We explain how each phase transition arises as a result of small nonzero eigenvalues in the Hessian matrix and demonstrate how this phenomenon is captured by susceptibilities.
\item We explain how the presence of linear features leads to an additional interpolation phase transition not present in an analogous model with nonlinear activation functions. We use random matrix theory to argue that the underlying reason for this difference is that nonlinear basis functions implicitly regularize small eigenvalues in the design matrix.
\end{itemize}

\section{Theoretical Setup}\label{sec:setup}

In this work, we focus on the supervised learning task of using relationships learned from a training data set, consisting of labels and associated input features, to accurately predict the labels of new data points from their input features.
Here, we closely follow the theoretical formalism previously described in Ref.~\onlinecite{Rocks2022}.

\subsection{Data Distribution (Teacher Model)}

We consider data points $(y, \vbx)$, each consisting of a continuous label $y$ paired with a set of $N_f$ continuous input features $\vbx$.
We assume that the relationship between the input features and labels (the data distribution or teacher model) can be expressed as 
\begin{align}
y(\vbx) &= y^*(\vbx; \vbbeta) + \varepsilon\label{eq:teacher}
\end{align}
where $\varepsilon$ is the label noise.
The unknown function $y^*(\vbx; \vbbeta)$ represents the ``true'' labels and depends on a set of $N_f$ ``ground truth'' parameters $\vbbeta$, characterizing the correlations between the features and labels.
Here, we restrict ourselves to a teacher model of the form
\begin{equation}
y^*(\vbx;\vbbeta) =  \frac{\sigma_\beta\sigma_X}{\expval*{f'}} f\qty(\frac{\vbx\cdot\vbbeta}{\sigma_X\sigma_\beta} ),\label{eq:true_labels}
\end{equation}
where the function $f$ is an arbitrary nonlinear function and ${\expval*{f'} = \frac{1}{\sqrt{2\pi}}\int_{-\infty}^\infty \dd he^{-\frac{h^2}{2}}f'(h)}$ is a normalization constant chosen for convenience with prime notation used to indicate a derivative.
Note that Eq.~\eqref{eq:true_labels} reduces to a linear teacher model  $y^*(\vbx) = \vbx\cdot\vbbeta$ when $f(h) = h$.

We draw the input features for each data point independently and identically from a normal distribution with zero mean and variance $\sigma_X^2/N_f$.
We consider ground truth parameters $\vbbeta$ and label noise $\varepsilon$ that are drawn independently from normal distributions with zero mean and variances $\sigma_\beta^2$ and $\sigma_\varepsilon^2$, respectively.
Furthermore, we assume the labels are centered so that $f$ has zero mean with respect to its argument.

\subsection{Model Architectures (Student Models)}

We consider a student model of the form
\begin{align}
\hat{y}(\vbx) &= \vbz(\vbx)\cdot\hbw,\label{eq:student}
\end{align}
where $\hbw$ is a vector of $N_p$ fit parameters. For the random linear features model, the vector of `hidden'' features $\vbz(\vbx)$ takes the form 
 \begin{align}
\vbz(\vbx) &= W^T\vbx,\label{eq:rlfr_model}
\end{align}
where $W$ is a random transformation matrix of size $N_f\times N_p$, whose elements are drawn independently from a normal distribution with zero mean and variance $\sigma_W^2/N_p$.

\subsection{Fitting Procedure}

We train each model on a training data set consisting of $M$ data points, ${\mathcal{D}=\{(y_a, \vbx_a)\}_{a=1}^M}$.
For convenience, we organize the vectors of input features in the training set into an observation matrix $X$ of size $M\times N_f$ and define the length-$M$ vectors of training labels $\vby$, training label noise $\vbeps$, and label predictions for the training set $\hby$.
We also organize the vectors of hidden features evaluated on the input features of the
training set, $\{\vbz(\vbx_a)  \}_{a=1}^M$, into the rows of a hidden feature matrix $Z$ of size $M\times N_p$.

Given a set of training data $\mathcal{D}$, we solve for the optimal values of the fit parameters $\hbw$ by minimizing the standard ridge regression loss function composed of the mean squared label error with $L_2$ regularization,
\begin{align}
L(\hbw; \mathcal{D}) &= \frac{1}{2}\norm*{\Delta \vby}^2 + \frac{\lambda}{2}\norm*{\hbw}^2,\label{eq:loss}
\end{align}
where the notation $\norm*{\cdot}$ indicates an $L_2$ norm,  ${\Delta \vby = \vby - \hby}$ is the vector of residual label errors for the training set, and $\lambda$ is the regularization parameter.
The exact solution for the fit parameters resulting from this loss function are
\begin{align}
\hbw &= \qty[\lambda I_{N_p} + Z^TZ]^{-1}Z^T\vby.\label{eq:exact}
\end{align}
We will often work in the ``ridge-less limit'' where we take the limit $\lambda \rightarrow 0$.
In this limit, we refer to the matrix $Z^TZ$ in the above expression as the Hessian matrix.

\subsection{Model Evaluation}

To evaluate a model's prediction accuracy, we measure the training and test (generalization) errors. 
We define the training error as the mean squared residual label error of the training data,
\begin{align}
\train  &= \frac{1}{M}\norm*{\Delta \vby}^2.
\end{align}
We define the interpolation threshold as the model complexity at which the training error becomes exactly zero (in the ridge-less limit). 
Analogously, we define the test error as the mean squared error evaluated on a test data set, ${\mathcal{D}'=\{(y_a', \vbx_a')\}_{a=1}^{M'}}$, composed of $M'$ new data points drawn independently from the same data distribution as the training set,
\begin{align}
\test &=  \frac{1}{M'}\norm*{\Delta \vby'}^2,\label{eq:test_err}
\end{align}
where $\Delta \vby' = \vby'-\hby'$ is a length-$M'$ vector of residual label errors between the vector of test labels $\vby'$ and their predicted values $\hby'$.
Furthermore, we define the ensemble-averaged training and test errors, $\etrain$ and $\etest$, respectively, by taking averages of the above definitions with respect to all sources of randomness (e.g., $X$, $\vbeps$, $\vbbeta$, etc.).

\subsection{Bias-Variance Decomposition}

The bias-variance decomposition separates test error into components stemming from three distinct sources: bias, variance, and noise. 
Here, we utilize the standard definitions of bias and variance~\cite{Geman1992, Bishop2006},
\begin{align}
\Bias\qty[\hat{y}(\vbx)] &= \E[\mathcal{D}]\qty[\hat{y}(\vbx)] - y^*(\vbx)\label{eq:raw_bias}\\
\Var\qty[\hat{y}(\vbx)] &= \E[\mathcal{D}]\qty[\hat{y}^2(\vbx)] - \E[\mathcal{D}]\qty[\hat{y}(\vbx)]^2,\label{eq:raw_var}
\end{align}
where the subscript $\mathcal{D}$ denotes the sampling average with respect to the training set (i.e., with respect to the input features $X$ and label noise $\vbeps$, but not the ground truth parameters $\vbbeta$).

In order to incorporate other sources of randomness (e.g., $\vbbeta$ and $W$), we define the more general ensemble-averaged squared bias and variance, respectively, as
\begin{align}
\expval*{\Bias^2[\hat{y}]} &= \E[\vbbeta, W, \vbx]\qty[\Bias[\hat{y}(\vbx)]^2]\\
\expval*{\Var[][\hat{y} ]} &=\E[\vbbeta, W, \vbx]\qty[\Var[][\hat{y}(\vbx)]] .
\end{align}
Using these definitions, we define the ensemble-averaged bias-variance decomposition of the test error,
\begin{align}
\etest &= \expval*{\Bias^2[\hat{y} ]} +\expval*{\Var[][\hat{y} ]} +\sigma_\varepsilon^2,\label{eq:bvdecomp}
\end{align}
which we utilize throughout this work.

\subsection{Derivation of Closed-Form Solutions}

Following the derivations in Ref.~\onlinecite{Rocks2022}, we utilize the zero-temperature cavity method to derive closed-form expressions for the training error, test error, bias, and variance.
In this derivation, we work in the thermodynamic limit, where $N_f, M, N_p \rightarrow \infty$, but their ratios, $\alpha_f = N_f/M$ and $\alpha_p = N_p / M$, remain finite.
Our results are exact in this limit.
Furthermore, we utilize the procedure described in Ref.~\onlinecite{Cui2020} to reproduce the closed-form solution for the eigenvalues spectrum of the Hessian matrix for this model.
We refer the reader to the Appendix for further details on these calculations.

\begin{figure*}[t!]
\centering
\includegraphics[width=1.0\linewidth]{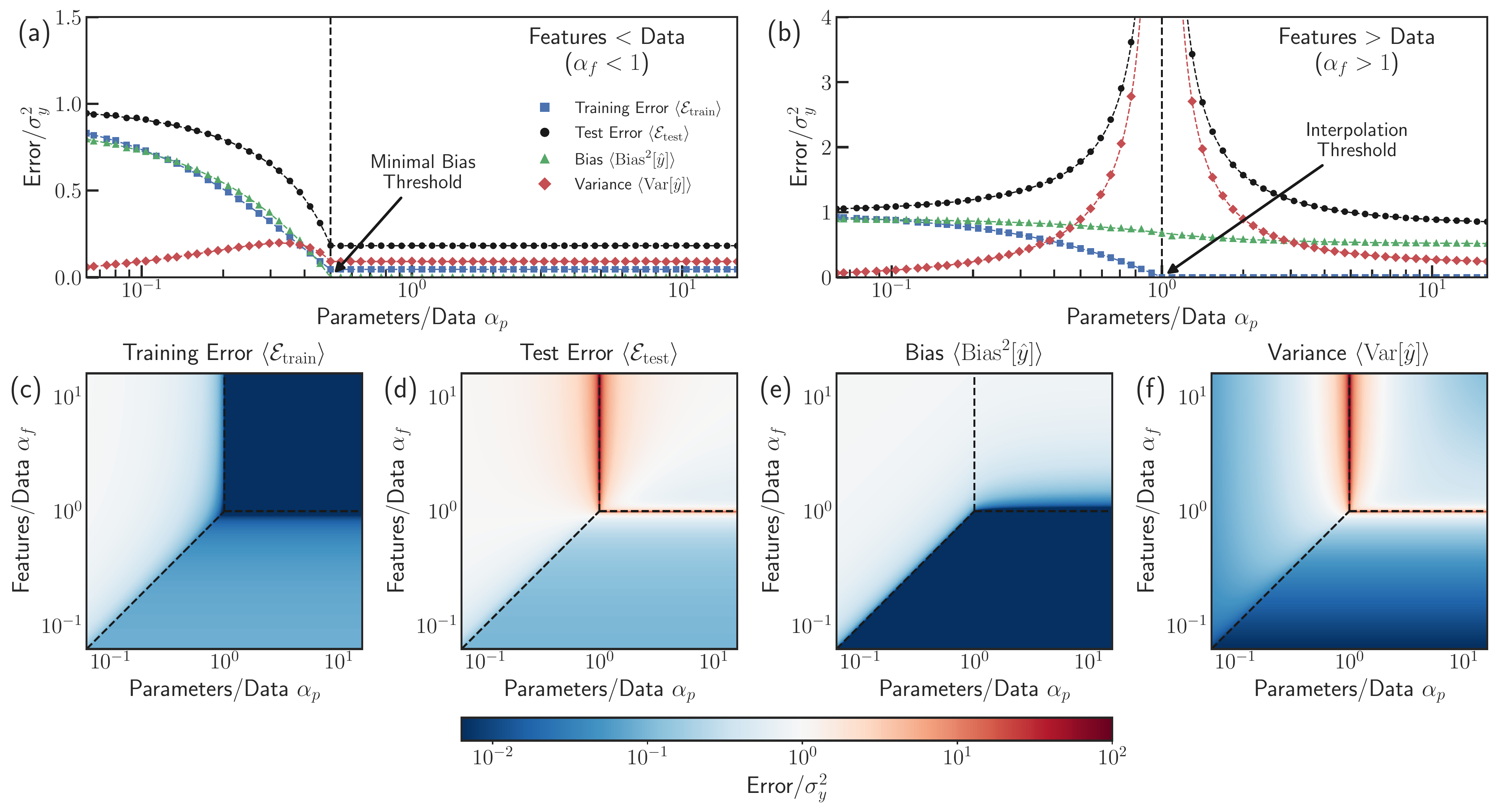} 
\caption{{\bf Random Linear Features Model  (Two-layer Linear Neural Network).}
Analytic solutions plotted as a function of $\alpha_p=N_p/M$ with fixed $\alpha_f=N_f/M$ for {\bf(a)} less input features than training data points ($\alpha_f=1/2$) and {\bf(b)} more input features than training data points ($\alpha_f=4$).
Shown are the ensemble-averaged training error (blue squares), test error (black circles), squared bias (green triangles), and variance (red diamonds).
Analytic solutions are indicated as dashed lines with numerical results shown as points. 
In (a), a black dashed vertical line marks the boundary between the large bias and minimal bias underparameterized regimes at $\alpha_p = \alpha_f$,
while in (b), a similar line marks the boundary between the under- and overparameterized regimes at $\alpha_p = 1$.
Analytic solutions as a function of $\alpha_p$ and $\alpha_f$ are also shown for the the ensemble-averaged {\bf(c)} training error, {\bf(d)} test error, {\bf(e)} squared bias, and {\bf(f)} variance.
Results are shown for a linear teacher model $y(\vbx) = \vbx\cdot\vbbeta + \varepsilon$, a signal-to-noise ratio of $\sigma_\beta^2\sigma_X^2 / \sigma_\varepsilon^2 = 10$, 
and have been scaled by the variance of the training set labels $\sigma_y^2 = \sigma_\beta^2\sigma_X^2 + \varepsilon^2$.
In each panel, black dashed lines indicate boundaries between different regimes of the solutions depending on which is the smallest of the quantities $M$, $N_f$, or $N_p$. 
See Appendix for additional numerical details.
}\label{fig:rlfm}
\end{figure*}

\section{Analytic Expressions}\label{sec:results}

We find that the closed-form solutions for the random linear features model are  characterized by three distinct regimes, each defined by which of the following three quantities is the smallest: the number of input features $N_f$, the number of fit parameters (hidden features) $N_p$, or the size of the training set $M$. 
In terms of $\alpha_f = N_f/M$ and $\alpha_p = N_p/M$, the expressions for ensemble-averaged the training error, test error, bias, and variance are
\begin{widetext}
\begin{alignat}{2}
\etrain &= \left\{
\begin{array}{c}
(\sigma_\varepsilon^2+\sigma_{\delta y^*}^2) (1-\alpha_f)\\
\sigma_\beta^2\sigma_X^2 \frac{(1-\alpha_p)(\alpha_f-\alpha_p)}{\alpha_f}  + (\sigma_\varepsilon^2+\sigma_{\delta y^*}^2) (1-\alpha_p)\\
0
\end{array}\right. &\quad & \begin{array}{l}
 \qif N_f < N_p, M\\
 \qif N_p < N_f, M\\
 \qif M < N_f, N_p
\end{array}
\label{eq:rlfm_train}\\
\etest &=  \left\{
\begin{array}{c}
(\sigma_\varepsilon^2+\sigma_{\delta y^*}^2)  \frac{1}{(1-\alpha_f)}\\
\sigma_\beta^2\sigma_X^2 \frac{(\alpha_f-\alpha_p)}{\alpha_f(1-\alpha_p)} +(\sigma_\varepsilon^2+\sigma_{\delta y^*}^2)  \frac{1}{(1-\alpha_p)}\\
\sigma_\beta^2\sigma_X^2 \frac{\alpha_p(\alpha_f-1)}{\alpha_f(\alpha_p-1)} + (\sigma_\varepsilon^2+\sigma_{\delta y^*}^2)  \frac{(\alpha_f\alpha_p-1)}{(\alpha_f-1)(\alpha_p-1)}
\end{array}
\right. &\quad & \begin{array}{l}
 \qif N_f < N_p, M\\
 \qif N_p < N_f, M\\
 \qif M < N_f, N_p
\end{array}\label{eq:rlfm_test}\\
\expval*{\Bias^2[ \hat{y}]} &= \left\{
\begin{array}{c}
\sigma_{\delta y^*}^2\\
 \sigma_\beta^2 \sigma_X^2\frac{(\alpha_f-\alpha_p)}{\alpha_f} + \sigma_{\delta y^*}^2 \\
 \sigma_\beta^2 \sigma_X^2\frac{\alpha_p(\alpha_f-1)^2}{\alpha_f(\alpha_f\alpha_p-1)} + \sigma_{\delta y^*}^2 
\end{array}
\right. &\quad & \begin{array}{l}
 \qif N_f < N_p, M\\
 \qif N_p < N_f, M\\
 \qif M < N_f, N_p
\end{array}\label{eq:rlfm_bias}\\
\expval*{\Var[][\hat{y}]} &=  \left\{
\begin{array}{c}
(\sigma_\varepsilon^2+\sigma_{\delta y^*}^2)  \frac{\alpha_f}{(1-\alpha_f)} \\
\sigma_\beta^2\sigma_X^2 \frac{\alpha_p(\alpha_f-\alpha_p)}{\alpha_f(1-\alpha_p)} + (\sigma_\varepsilon^2+\sigma_{\delta y^*}^2) \frac{\alpha_p}{(1-\alpha_p)} \\
\sigma_\beta^2\sigma_X^2 \frac{\alpha_p(\alpha_f-1)(\alpha_f-1+\alpha_p-1)}{\alpha_f(\alpha_p-1)(\alpha_f\alpha_p-1)} + (\sigma_\varepsilon^2+\sigma_{\delta y^*}^2)   \frac{(\alpha_f-1+\alpha_p-1)}{(\alpha_f-1)(\alpha_p-1)} 
\end{array}
\right. &\quad & \begin{array}{l}
 \qif N_f < N_p, M\\
 \qif N_p < N_f, M\\
 \qif M < N_f, N_p,
\end{array}\label{eq:rlfm_var}
\end{alignat}
\end{widetext}
where we have taken the limit $\lambda \rightarrow 0$ (with leading order terms of order $\lambda^2$ reported in the Appendix for quantities reported here as zero).
The quantity $\sigma_{\delta y^*}^2$ is the statistical variance of the nonlinear components of the true labels, measured via their deviation from a linear teacher model, $\sigma_{\delta y^*}^2 = \E[\vbx][(y^*(\vbx) - \vbx\cdot\vbbeta)^2]$.
In the thermodynamic limit, we find that this quantity evaluates to
\begin{equation}
\begin{gathered}
\sigma_{\delta y^*}^2 = \sigma_\beta^2\sigma_X^2 \Delta f\qqc \Delta f = \frac{\expval*{f^2}-\expval*{f'}^2}{\expval*{f'}^2}\\
\expval*{f^2} = \frac{1}{\sqrt{2\pi}}\int\limits_{-\infty}^\infty \dd h e^{-\frac{1}{2}h^2} f^2(h)\\
\expval*{f'} = \frac{1}{\sqrt{2\pi}}\int\limits_{-\infty}^\infty \dd h e^{-\frac{1}{2}h^2} f'(h).
\end{gathered}
\end{equation}

In Figs.~\ref{fig:rlfm}(a) and (b), we plot the training error, test error, bias, and variance as a function of $\alpha_p=N_p/M$ for fixed $\alpha_f=N_f/M$ for the two cases $\alpha_f < 1$ and $\alpha_f >1$, respectively.
To gain a better grasp of the full set of solutions, 
we also plot all quantities in Eqs.~\eqref{eq:rlfm_train}-\eqref{eq:rlfm_var} as a function of both $\alpha_p$ and $\alpha_f$ in Figs.~\ref{fig:rlfm}(c)-(f).
All solutions are  shown for a linear teacher model ${y^*(\vbx) = \vbx\cdot\vbbeta}$  (${\sigma_{\delta y^*}^2 = 0}$).

The three regimes we observe in the closed-form solutions are separated by three distinct phase transitions.
Examining the training error in Fig.~\ref{fig:rlfm}(c), 
we find that it goes to zero at two of the transitions, $\alpha_p = 1$ with $\alpha_f \geq 1$ and $\alpha_f = 1$ with $\alpha_p \geq 1$, giving rise to an interpolation boundary.
On one side of this boundary, where $\alpha_p < 1$ or $\alpha_f < 1$ (there are less data points $M$ than fit parameters $N_p$ or input features $N_f$), the model is underparameterized, while beyond this boundary the model is overparameterized and in the ``interpolation'' regime.
We note that the interpolation transition for this model is markedly different from that of the random nonlinear features model (nonlinear activation function) where the interpolation threshold occurs at $\alpha_p = 1$ independently of $\alpha_f$ (see Sec.~\ref{sec:compare} and Ref.~\onlinecite{Rocks2022}).

We find that the test error and variance diverge along the entire interpolation boundary, while the bias remains finite. 
This is in contrast with previous studies which employed non-standard definitions of bias and variance and found that both variance and bias diverge at the interpolation threshold~\cite{Ba2020}.
Examining Figs.~\ref{fig:rlfm}(a) and (b), we find that the test error (and similarly, the variance) exhibits very different behavior as a function of $\alpha_p$, 
depending on whether $\alpha_f < 1$ (less input features than training data points $N_f < M$) or $\alpha_f >1$ (more input features than training data points $N_f>M$).
When $\alpha_f > 1$, the test error diverges at $\alpha_p=1$ and decreases monotonically in the overparameterized regime.
In contrast, when $\alpha_f < 1$, the test error monotonically decreases to a small, constant value at $\alpha_p \geq \alpha_f$.
Although this model does not display the full canonical double-descent behavior in either case, the test error in the 
overparameterized regime is always at least as small as -- if not smaller than -- that of the underparameterized regime for fixed $\alpha_f$.

Examining the bias in Fig~\ref{fig:rlfm}(c) reveals that there is an additional phase transition in the underparameterized regime located at the boundary $\alpha_f=\alpha_p$  for $\alpha_p \leq 1$ and $\alpha_f \leq 1$  (i.e., when the number of input features $N_f$ equals the number of hidden features $N_p$, with both $N_f$ and $N_p$ less than the number of data points $M$). 
This transition divides the non-interpolation solutions into two pieces.
When $\alpha_p < \alpha_f$ (less fit parameters than input features $N_p < N_f$),
the model exhibits a large bias because there are not enough fit parameters (or hidden features) to fully express the input features in the data [see Eq.~\eqref{eq:rlfm_bias}].
In contrast, when $\alpha_p > \alpha_f$ (more fit parameters than input features $N_p > N_f$), the only contribution to the bias is a small constant $\sigma_{\delta y^*}^2$ stemming from the nonlinear components of the labels.
For the special case of a linear teacher model shown in the figures, the bias is identically zero in this regime.

Interestingly, we also observe that $\sigma_{\delta y^*}^2$ appears as an additive component to the label noise $\sigma_\varepsilon^2$ in the training error, test error, and variance,
indicating that the model interprets the nonlinear components of the labels as effective noise~\cite{Rocks2022}.

\begin{figure*}[t!]
\centering
\includegraphics[width=\linewidth]{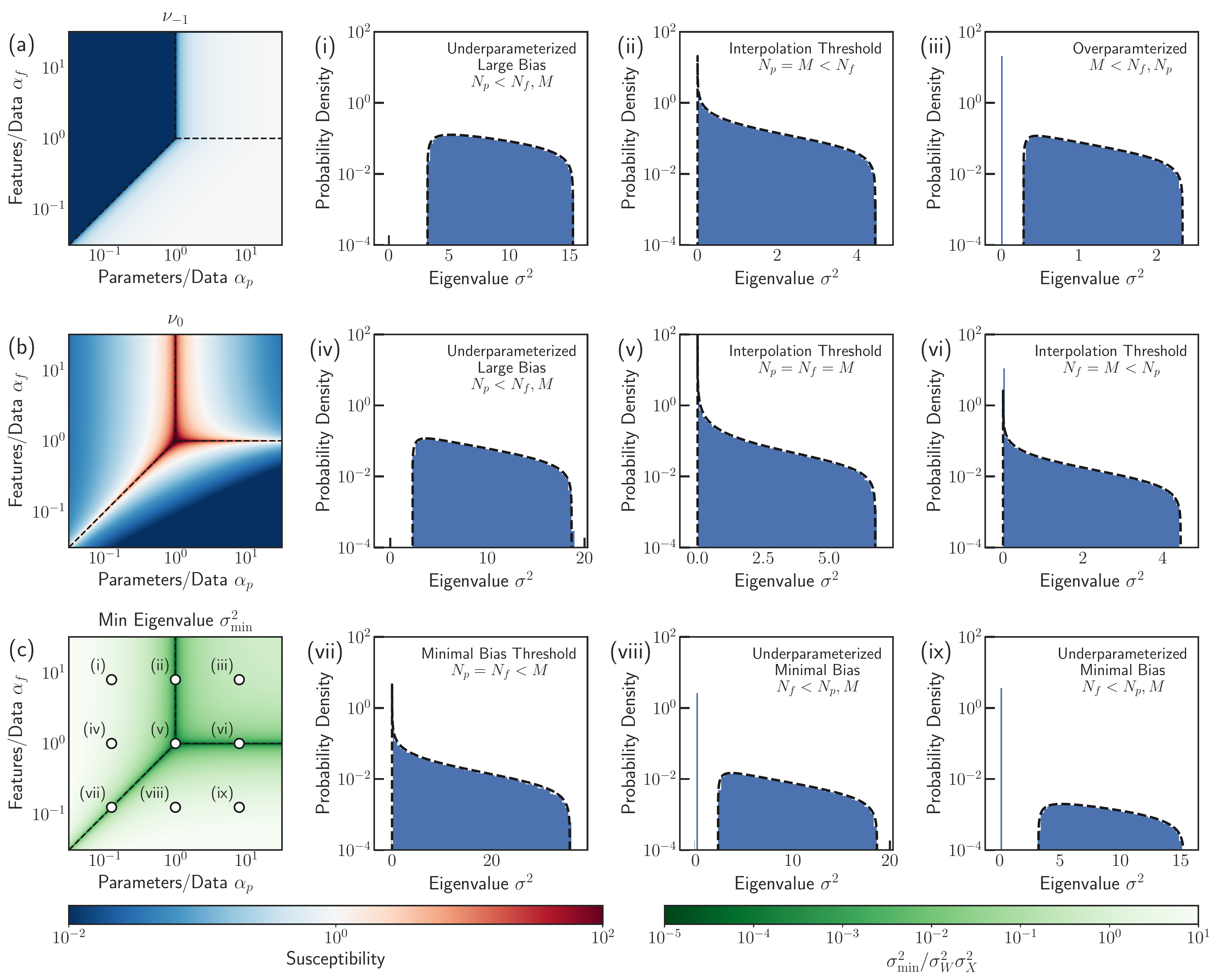} 
\caption{
{\bf Susceptibilities and Eigenvalue Spectra}
{\bf(a)-(b)} Analytic solutions for the susceptibility $\nu$ which measures the sensitivity of the fit parameters to small perturbations in the gradient. 
In the small $\lambda$ limit,  ${\nu \approx \lambda^{-1}\nu_{-1} + \nu_0}$.
{\bf(a)} The coefficient $\nu_{-1}$ counts the fraction of unutilized fit parameters, or the fraction of parameters beyond that needed to attain minimal training error.
{\bf(b)} The coefficient $\nu_0$ diverges at each phase transition when $Z^TZ$ has a small eigenvalue.
{\bf(c)} Analytic solution for the minimum nonzero eigenvalue $\sigma_{\min}^2$ of the Hessian matrix $Z^TZ$. 
{\bf(i)-(ix)} Examples of the full eigenvalue spectrum are shown for each of the corresponding points in (c).
See Appendix for additional simulation details.
}\label{fig:spectrum}
\end{figure*}

\section{Phase transitions, susceptiblities, and eigenvalue spectra}\label{sec:spectra}

In the previous section, we found that the analytic solutions for the random linear features model are characterized by three distinct regimes separated by three different phase transitions.
As a natural byproduct of our cavity derivations, we find that each of these phase transitions is marked by a diverging susceptibility.
In particular, setting the gradient of the loss function in Eq.~\eqref{eq:loss} equal to a small nonzero field $\vbeta$, such that $\pdv*{L}{\hbw} = \vbeta$, we define the susceptibility matrix $\pdv*{\hbw}{\vbeta}$.
This quantity measures  the sensitivity of the fit parameters to small perturbation in the gradient and can be shown to be equivalent to the inverse Hessian of the loss function.
Taking the trace of this matrix, we define the scalar susceptibility
\begin{equation}
\nu = \frac{1}{N_p}\Tr \pdv{\hbw}{\vbeta} = \frac{1}{N_p}\Tr \qty[\lambda I_{N_p} + Z^TZ]^{-1}.
\end{equation}
In the small $\lambda$ limit, we make the approximation ${\nu \approx \lambda^{-1} \nu_{-1} + \nu_0}$. In exact matrix form, we find that the two coefficients are
\begin{align}
\nu_{-1} = 1 - \frac{1}{N_p}\rank(Z^TZ)\qc \nu_0 = \frac{1}{N_p}\Tr [Z^TZ]^+,\label{eq:nucoeffs}
\end{align}
where $^+$ denotes a Moore-Penrose pseudoinverse.

In Figs.~\ref{fig:spectrum}(a) and (b), we plot the analytic closed-form expressions for these two quantities in the thermodynamic limit  as a function of $\alpha_f$ and $\alpha_p$ (see Appendix for expressions).
We find that the first coefficient $\nu_{-1}$ counts the fraction of fit parameters that go beyond the minimum needed to attain minimal training error.
In contrast, the second coefficient $\nu_0$ diverges along each phase boundary.
Based on the exact matrix form of $\nu_0$ in Eq.~\eqref{eq:nucoeffs}, we infer that these divergences can be attributed to small eigenvalues in the Hessian matrix $Z^TZ$.

To illustrate this connection between the eigenvalues of the Hessian and the susceptibility $\nu$, we note that for this problem, the inverse Hessian is equivalent to the Green's function  and can be used to extract the eigenvalue spectrum~\cite{Cui2020} (see Appendix for derivation). 
In Fig.~\ref{fig:spectrum}(c), we show the analytic solution for the minimum nonzero eigenvalue $\sigma_{\min}^2$ of $Z^TZ$.
Consistent with $\nu_0$, we find that $\sigma_{\min}^2$ goes to zero along each phase transition.
In Figs.~\ref{fig:spectrum}(i)-(ix), we also plot the eigenvalue distributions for the points indicated in Fig.~\ref{fig:spectrum}(c).
While $\nu_0$ captures the distribution of nonzero eigenvalues, $\nu_{-1}$ captures the weight of the delta function at zero in the overparameterized and minimal bias regimes.
In each regime and along each phase boundary, these distributions are qualitatively similar to the Marchenko-Pastur distribution~\cite{Marchenko1967}.
Along each phase transition, the gap in the distribution goes to zero, while the gap is finite away from each boundary.
The presence of this eigenvalue gap was previously shown to be the root cause of the decrease in variance in the overparameterized regime~\cite{Rocks2022}.

To understand the source of these small eigenvalue gaps,
we note that the types of random matrices we consider in this work typically exhibit infinitesimally small eigenvalues in the thermodynamic limit if they contain an equal number of rows and columns.
This fact allows us to identify which matrix is the root cause of each transition.
Since $Z$ is a product of $X$ and $W$, this phenomenon arises in two forms.
First, $Z$ can exhibit a small eigenvalue if either $X$ or $W$ is square and the expression of its input feature space is not limited by its product with the other matrix (e.g., if $X$ is square and $N_p \geq N_f = M$ or $W$ is square and $M \geq N_f = N_p$).
This behavior explains the interpolation transition at $\alpha_f = 1$ which arises due to small eigenvalues in $X$, but does not extend below $\alpha_p = 1$ when the rank of $W$ becomes too low to preserve every direction in the space of input features encoded in $X$.
Similarly, the minimal bias transition at $\alpha_f = \alpha_p$ occurs due to small eigenvalues in $W$, disappearing above $\alpha_f  = 1$ when the rank of $X$ is too low to fully express the input feature space of $W$.
Second, $Z$ can exhibit a small eigenvalue if it is square and full rank, giving rise to the transition at $\alpha_p = 1$, but only when $\alpha_f \geq 1$.

Finally, we observe that the test error only diverges at the two phase transitions to the interpolation regime, but not at the minimal bias transition, despite $\sigma_{\min}^2$ going to zero in all three cases.
This lack of divergence is explained by the fact that the minimal bias transition arises due to small eigenvalues in $W$, which is used to transform both the training data and the test data.
Since both data sets are transformed in the same way, predictions by the model for the test set based on the training set will not be limited by small eigenvalues in $W$, and the test error will not diverge when $W$ is square ($N_f = N_p$).

\begin{figure*}[t!]
\centering
\includegraphics[width=\linewidth]{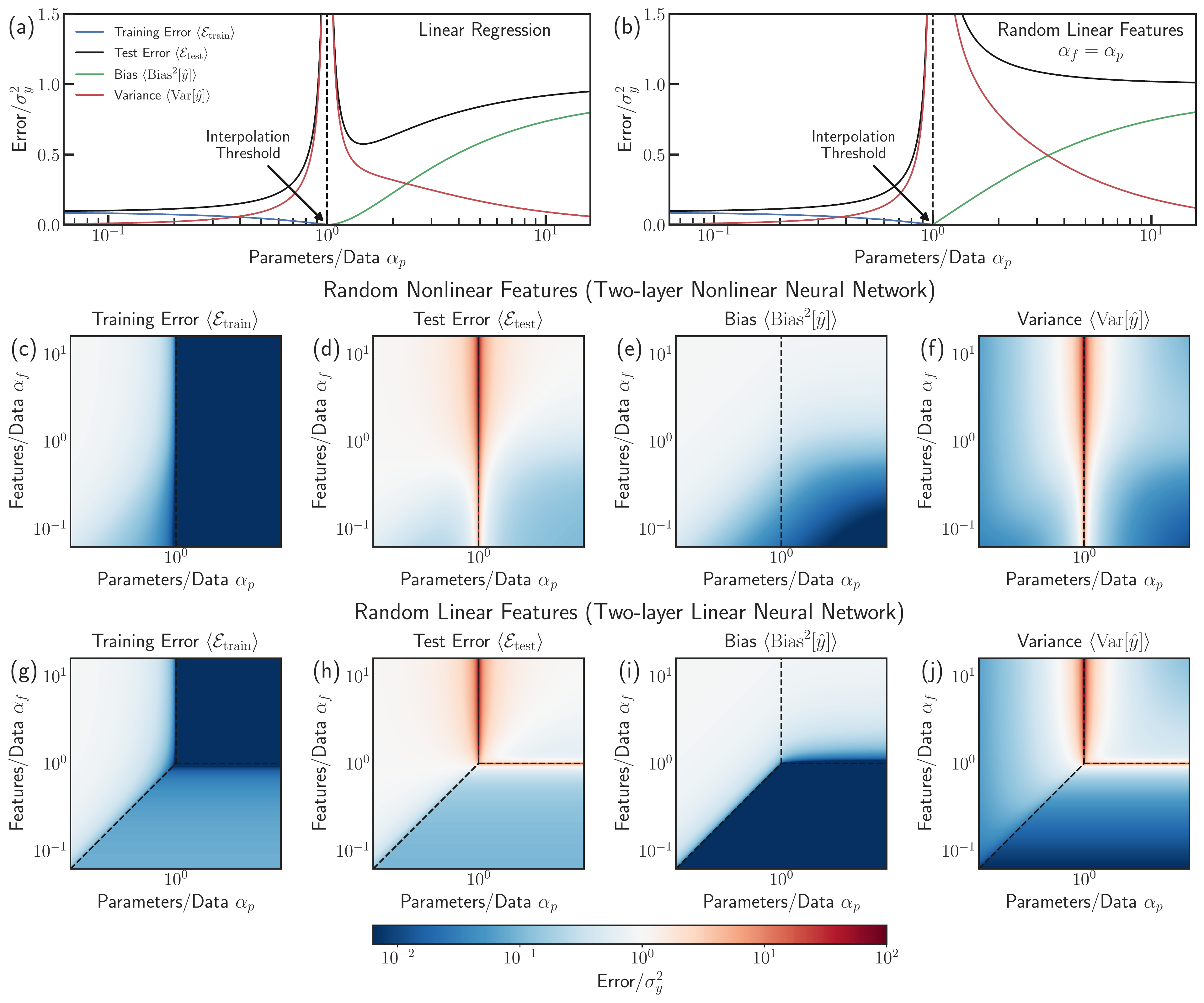} 
\caption{
{\bf Comparison of the Random Linear Features Model to Linear Regression (No Basis Functions) and the Random Nonlinear Features Model.}
{\bf(a)} The training error, test error, bias, and variance as a function of $\alpha_p$ (or equivalently, $\alpha_f$) for linear regression (no basis functions) ,
and {\bf(b)} the same quantities for the random linear features model for the special case $\alpha_f = \alpha_p$.
  The training error, test error, bias, and variance as a function of $\alpha_p$ and $\alpha_f$ for {\bf(c)-(f)} the random nonlinear features model with ReLU activation, $\varphi(h) = \max(0, h)$, and 
{\bf(g)-(j)} the  random linear features model. In all panels, black dashed lines indicate phase transitions.
Results for linear regression and the random nonlinear features model are reproduced from Ref.~\onlinecite{Rocks2022}.
}\label{fig:compare}
\end{figure*}

\section{Comparison to Linear Regression and the Random Nonlinear Features Model}\label{sec:compare}

One of the more surprising results of our analysis is that the phase diagram for the random \emph{linear} features model is qualitatively different from the random \emph{nonlinear} features model.
On other hand, we find that the random linear feature model is qualitatively similar to ordinary ridge-less regression when the number of hidden features matches the input features.
To better understand the similarities and differences, we have reproduced the phase diagrams for all three models in Fig.~\ref{fig:compare} (see Ref.~\cite{Rocks2022} for a detailed analysis of ridge regression and the random nonlinear features model).

First, we compare the random linear features model to linear regression in which the number of hidden features matches the input features,
\begin{equation}
\vbz(\vbx) = \vbx.
\end{equation}
Fig.~\ref{fig:compare}(a) shows the training error, test error, bias, and variance for linear regression.
Since linear regression lacks basis functions, in Fig.~\ref{fig:compare}(b), we plot the same quantities for the random linear features model for the special case where the number of input features equals the number of hidden features ($\alpha_f = \alpha_p$).

We observe that along this cut of the phase diagram, the random linear features model behaves qualitatively similar to linear regression, with variance first increasing as one approaches the interpolation threshold ($\alpha_p=1$) and then decreasing monotonically beyond the threshold. 
Meanwhile, the bias is zero below the interpolation threshold and then increases once one crosses the interpolation threshold. 
As discussed in detail in Ref.~\cite{Rocks2022}, the underlying reason for the increase in bias for $\alpha_p>1$ is that in this regime, the model does not have enough training data points to sample the entire input feature space. 
Therefore, any predictions made about these unsampled directions represent implicit assumptions of the model.

Next, we compare to the random nonlinear features model in which the hidden features take the form
\begin{equation}
\vbz(\vbx) = \frac{1}{\expval*{\varphi'}}\frac{\sigma_W\sigma_X}{\sqrt{N_p}}\varphi\qty(\frac{\sqrt{N_p}}{\sigma_W\sigma_X}W^T\vbx)
\end{equation}
where $\varphi$ is a nonlinear activation function that acts separately on each element of its input and ${\expval*{\varphi'} = \frac{1}{\sqrt{2\pi}}\int\limits_{-\infty}^\infty dhe^{-\frac{h^2}{2}}\varphi'(h)}$ is a normalization constant.
Figs.~\ref{fig:spectrum}(c)-(f) show the training error, test error, bias, and variance for this model for the case of ReLU activation, $\varphi(h) = \max(0, h)$, as a function of both $\alpha_p$ and $\alpha_f$, while Figs.~\ref{fig:spectrum}(g)-(j) depict the same for the random linear features model.

We observe that while the interpolation transition boundary at $\alpha_p=1$ for $\alpha_f \geq 1$ remains the same, the addition of a nonlinear activation function suppresses the interpolation transition at $\alpha_f = 1$ for $\alpha_p > 1$, along with the transition to a minimal bias regime at $\alpha_f=\alpha_p$  for $\alpha_p < 1$.
At the same time, the interpolation transition at $\alpha_p=1$ is extended to all values of $\alpha_f$.

The two changes to the shape of the interpolation boundary can be attributed to the behavior of small eigenvalues in the Hessian.
Upon the introduction of a nonlinear activation, the interpolation transition at $\alpha_p = 1$  for $\alpha_f< 1$ arises due to the creation of new small eigenvalues in $Z$.
The nonlinear transformation promotes $Z$ to full rank when it is square ($N_p = M$), even if the product $XW$ is not full rank.
As a result, $Z$ exhibits small eigenvalues at this transition whether or not $XW$ has a small eigenvalue, translating to small eigenvalues in the Hessian and a divergence in the test error.
In contrast, our random matrix theory analysis suggests that the presence of a nonlinear activation function suppresses the transition at $\alpha_f=1$ with $\alpha_p > 1$ by serving as an implicit regularizer of small eigenvalues in the Hessian (this behavior was previously observed in Ref.~\onlinecite{DAscoli2020}). 
In particular, the use of nonlinear basis functions masks divergences arising from small eigenvalues that arise when the design matrix $X$ is square ($N_f = M$). 

Finally, to account for the removal of the minimal bias transition when $\alpha_f=\alpha_p$ with $\alpha_p < 1$, we note that the training data is generated using a teacher model that depends directly on the input features, while the nonlinear model first applies a nonlinear transformation.
This nonlinear basis masks properties of the underlying input feature space like its dimension, introducing additional bias.
Therefore, the bias does not approach a minimal value at $\alpha_p = \alpha_f$, even if there are in principle enough hidden features to fully encode the full space of input features.
Instead, we observe that the bias only reaches a minimum in the limit $\alpha_p \rightarrow \infty$ for fixed $\alpha_f$.

\section{Conclusions}\label{sec:conclude}

A central question in machine learning is understanding why complicated models with many more parameters than training data points can still make accurate predictions. Here, we have tackled this problem by analyzing one of the simplest examples of non-trivial supervised learning: regression with random linear features. Despite the simplicity of the model, it exhibits remarkably rich behavior with multiple phase transitions. 

We found that the phase diagram of the model has three distinct phases: (i) an underparameterized regime with finite training error and large bias, (ii) a second underparameterized regime with minimal bias, and (iii) an overparameterized, or interpolation, regime with zero training error. We also showed that at the transition to the interpolation regime, the variance but not the bias diverges. For this reason, while the classical bias-variance trade-off captures much of the behavior of the model in the underparameterized regime, it fails to describe the interpolation regime where the variance decreases with increasing model complexity. 

We showed that the divergence of the variance is due to the presence of small eigenvalues in the Hessian matrix. This is consistent with the general picture advocated in Refs.~\onlinecite{Advani2020, Rocks2022} that large test errors are associated with the closing of a spectral gap near the interpolation transition. On both sides of the transition, when the spectral gap is large, it is easy to distinguish noise from poorly sampled directions in feature space. However, when the gap closes this is no longer possible, accounting for the large variance. 

Our work suggests that many of the fundamental features of double-descent can be understood even when considering simple convex models. An important question is how to generalize the intuitions developed here to more complex settings.  In contrast with the random linear features model, modern deep learning methods are often non-convex and attempt to learn meaningful features directly from data.  In the future, it will be interesting to see how this changes the understanding of the bias-variance trade-off developed here~\cite{Chaudhari2019, Baldassi2020, Pittorino2020}. 

\section*{Acknowledgments}

This work was supported by NIH NIGMS grant 1R35GM119461 and a Simons Investigator in the Mathematical Modeling of Living Systems (MMLS) award to PM. 
The authors also acknowledge support from the Shared Computing Cluster administered by Boston University Research Computing Services.


\input{manuscript_rlfm_bias_var.bbl}

\onecolumngrid

\appendix

\section{Cavity Derivations}

In this section, we provide detailed derivations of all closed-form solutions for the random linear features model. 
These calculations follow the general procedure laid out in Ref.~\onlinecite{Rocks2022}.

\subsection{Notational Conventions}

\begin{itemize}
\item We define $M$ as the number of points in the training data set, $N_f$ as the number of input features, and $N_p$ as the number of fit parameters/hidden features. We define the ratios $\alpha_f = N_f/M$ and $\alpha_p = N_p/M$.
\item Unless otherwise specified, the type of symbol used for an index label (e.g., $\Delta y_a$) or as a summation index (e.g., $\sum_a$) implies its range. The symbols $a$, $b$, or $c$ imply ranges over the training data points from $1$ to $M$, 
the symbols $j$, $k$, or $l$ imply ranges over the input features from $1$ to $N_f$, and the symbols $J$, $K$, or $L$ imply ranges over the fit parameters/hidden features from $1$ to $N_p$.
\item The notation $\E[x]\qty[\cdot]$, $\Var[x]\qty[\cdot]$ and $\Cov[x]\qty[\cdot, \cdot]$ represent the mean, variance, and covariance, respectively, with respect to one or more random variables $x$. 
A lack of subscript implies averages taken with respect to the total ensemble distribution, i.e., taken over all possible sources of randomness.
A subscript $0$ implies averages taken with respect to random variables containing one or more $0$-valued indices (e.g., $X_{a0}$, $X_{0j}$, $W_{0J}$, or $W_{j0}$).
\end{itemize}

\subsection{Nonlinear Label Decomposition}

In order to calculate the statistical properties of the nonlinear teacher model, 
we first decompose the labels into their linear and nonlinear components as follows:
\begin{equation}
y(\vbx) = \vbx\cdot\vbbeta + \delta y^*_{\mathrm{NL}}(\vbx) + \varepsilon\qqc\vbbeta \equiv \Sigma_{\vbx}^{-1}\Cov[\vbx][\vbx, y^*(\vbx)].\label{eq:SIydecomp}
\end{equation}
The first term in this decomposition captures the linear correlations between the labels $y^*$ and the input features $\vbx$ via the ground truth parameters $\vbbeta$.
The second term, defined as ${\delta y^*_{\mathrm{NL}}(\vbx)\equiv y^*(\vbx) -  \vbx\cdot\vbbeta}$, represents the remaining nonlinear component of the labels.
By defining the ground truth parameters as shown above, where $\Sigma_{\vbx} = \Cov[\vbx][\vbx, \vbx^T]$ is the covariance matrix of the input features (assumed to be invertible), it can be proven that the linear and nonlinear components are statistically independent. Using the definition of the nonlinear component of the labels $\delta y^*_{\mathrm{NL}}(\vbx)$, it is straightforward to show that their mean and covariance in the thermodynamic limit evaluate to
\begin{equation}
\E\qty[\delta y^*_{\mathrm{NL}}(\vbx_a)] = 0\qqc\Cov\qty[\delta y^*_{\mathrm{NL}}(\vbx_a), \delta y^*_{\mathrm{NL}}(\vbx_b)] = \sigma_{\delta y^*}^2\delta_{ab},
\end{equation}
where $\vbx_a$ and $\vbx_b$ are to independent data points and we have defined the variance $\sigma_{\delta y^*}^2$ of the nonlinear components as
\begin{equation}
\begin{gathered}
\sigma_{\delta y^*}^2 = \sigma_\beta^2\sigma_X^2 \Delta f\qqc \Delta f = \frac{\expval*{f^2} -\expval*{f'}^2 }{\expval*{f'}^2}\\
\expval*{f^2} = \frac{1}{\sqrt{2\pi}}\int\limits_{-\infty}^\infty dh e^{-\frac{h^2}{2}}f^2(h)\qqc\expval*{f'} = \frac{1}{\sqrt{2\pi}}\int\limits_{-\infty}^\infty dh e^{-\frac{h^2}{2}}f'(h).
\end{gathered}
\end{equation}
The decomposition in Eq.~\eqref{eq:SIydecomp} and its statistical properties are derived in detail in Ref.~\onlinecite{Rocks2022}.

\subsection{General Solutions}\label{sec:gen_sol}

Next, we derive some useful formulas for the ensemble-averaged quantities we wish to calculate.
First, we express the ensemble-averaged training error as
\begin{equation}
\etrain = \expval*{\Delta y^2}\qqc \expval*{\Delta y^2} = \E\qty[\frac{1}{M}\sum_b \Delta y_b^2]
\end{equation}
where we have defined $\expval*{\Delta y^2}$ as the mean squared label error for the training data. 

Next, we evaluate the average over the test data set in the ensemble average of the test error,
\begin{equation}
\etest = \sigma_X^2 \expval*{\Delta\beta^2} + \sigma_{\delta y^*}^2 + \sigma_\varepsilon^2.
\end{equation}
To obtain this expression, we have defined the set of ground truth parameters estimated by the model as $\hbbeta \equiv W\hbw$ and the corresponding residual parameter error $\Delta \vbbeta \equiv \vbbeta - \hbbeta$. The quantity $\expval*{\Delta\beta^2}$ is then the mean squared residual parameter error,
\begin{equation}
\expval*{\Delta\beta^2} = \E\qty[\frac{1}{N_f}\sum_k \Delta \beta_k^2].
\end{equation}

To correctly calculate the ensemble average of the squared bias, we make use of the following trick: we reinterpret the square of the average over $\mathcal{D}$ as two separate averages over uncorrelated training data sets,
\begin{equation}
\begin{aligned}
\Bias^2[\hat{y}(\vbx)] &= (\E[\mathcal{D}]\qty[\hat{y}(\vbx)] - y^*(\vbx))^2\\
&= \E[\mathcal{D}_1, \mathcal{D}_2]\qty[\qty(y(\vbx) - \hat{y}_1(\vbx))\qty(y(\vbx) - \hat{y}_2(\vbx))].\label{eq:biastwodatasets}
\end{aligned}
\end{equation}
Now, instead of a single regression problem trained on a single data set $\mathcal{D}$, we consider two separate regression problems each trained independently on different training sets, $\mathcal{D}_1$ and $\mathcal{D}_2$, drawn from the same distribution with the same ground truth parameters $\vbbeta$.
These regression problems will also share all other random variables including the test data point $(y, \vbx)$, $W$, etc.

Next, we apply the ensemble average and explicitly average over the test data point $\vbx$ to obtain
\begin{equation}
\expval*{\Bias^2[\hat{y}(\vbx)]} = \sigma_X^2 \expval*{\Delta\beta_1\Delta\beta_2}\qqc\expval*{\Delta\beta_1\Delta\beta_2} = \E\qty[\frac{1}{N_f}\sum_k \Delta \beta_{1, k}\Delta_{2,k}]\label{eq:SIgen_bias}
\end{equation}
where we have defined $\expval*{\Delta\beta_1\Delta\beta_2}$ as the covariance of the residual label errors between the two models trained on data sets $\mathcal{D}_1$ and $\mathcal{D}_2$.

Finally, we find an expression for the variance by subtracting the bias and noise ($\sigma_\varepsilon^2$) from the test error,
\begin{equation}
\expval*{\Var[][\hat{y}(\vbx)]}  = \sigma_X^2  \qty(\expval*{\Delta\beta^2}- \expval*{\Delta\beta_1\Delta\beta_2}).\label{eq:SIgen_var}
\end{equation}

Based on these expressions, we find that the training error, test error, bias, and variance depend on three key ensemble-averaged quantities: $\expval*{\Delta y^2}$, $\expval*{\Delta \beta^2}$, and $\expval*{\Delta\beta_1\Delta\beta_2}$.
We aim to calculate these quantities in the remainder of this derivation.

\subsection{Linear System of Equations}

In this section, we derive a linear system of equations to which we will apply the cavity method.
To do this, we first evaluate the gradient of the loss function in Eq.~\eqref{eq:loss} with respect to the fit parameters,
\begin{equation}
0 = \pdv{L (\hbw)}{\hat{w}_J} = -\sum_b\Delta y_b Z_{bJ} + \lambda \hat{w}_J.\label{eq:system}
\end{equation}
In addition to this gradient equation, we will also need the equations for the residual label errors for the training set,
\begin{equation}
\Delta y_a = y^*(\vbx_a)  + \varepsilon_a -  \sum_K \hat{w}_K Z_{aJ}.
\end{equation}

Next, we decompose these two sets of equations such that they are linear in the random matrices $W$ and $X$, resulting in four different sets of equations,
\begin{equation}
\begin{aligned}
\lambda \hat{w}_J &= \sum_k \hat{u}_k  W_{kJ} + \eta_J\\
\hat{u}_j &= \sum_b\Delta y_b X_{b j}  + \psi_j\\
\Delta y_a &=  \sum_k \Delta \beta_k X_{a k}+ \delta y^*_{\mathrm{NL}}(\vbx_a) + \varepsilon_a + \xi_a\\
\Delta \beta_j &= \beta_j - \sum_K \hat{w}_K W_{jK}+ \zeta_j,
\end{aligned}
\end{equation}
where we have also utilized Eq.~\eqref{eq:SIydecomp} to decompose the training labels into their linear and nonlinear components.
We have also added a small auxiliary field, $\eta_J$, $\psi_j$, $\xi_a$, or $\zeta_j$, to each equation.
We will use these extra fields to define perturbations about the solutions to these equations with the intent of setting the fields to zero at the end of the derivation.
The quantities $\hat{u}_j$ can be interpreted as representations of the fit parameters in the space of input features.

\subsection{Cavity Expansion}

Next, we add an additional variable of each type, resulting in a total of $M+1$ data points, $N_f+1$ input features and $N_p+1$ fit parameters. 
Each additional variable is represented using an index value of $0$, written as $\hat{w}_0$, $\hat{u}_0$, $\Delta y_0$, and $\Delta \beta_0$.
After including these new unknown quantities, the four equations become
\begin{equation}
\begin{aligned}
\lambda \hat{w}_J &= \sum_k  \hat{u}_k W_{kJ} + \eta_J +  \hat{u}_0 W_{0J}\\
\hat{u}_j &= \sum_b \Delta y_b  X_{b j} + \psi_j + \Delta y_0  X_{0j}\\
\Delta y_a &=  \sum_k \Delta \beta_kX_{a k} + \delta y^*_{\mathrm{NL}}(\vbx_a)  + \varepsilon_a + \xi_a +  X_{a 0}\Delta \beta_0\\
\Delta \beta_j &= \beta_j - \sum_K \hat{w}_KW_{jK} + \zeta_j -  \hat{w}_0W_{j0},\label{eq:rlfm_plus0}
\end{aligned}
\end{equation}
with each new variable described by a new equation,
\begin{equation}
\begin{aligned}
\lambda \hat{w}_0 &= \sum_k  \hat{u}_k W_{k0} + \eta_0 +  \hat{u}_0 W_{00}\\
\hat{u}_0 &= \sum_b \Delta y_b  X_{b 0} + \psi_0 +  \Delta y_0  X_{0 0}\\
\Delta y_0 &=  \sum_k \Delta \beta_kX_{0 k} +  \delta y^*_{\mathrm{NL}}(\vbx_0) + \varepsilon_0 + \xi_0 + \Delta \beta_0 X_{0 0}\\
\Delta \beta_0 &= \beta_0 - \sum_K\hat{w}_K W_{0K} + \zeta_0 -   \hat{w}_0 W_{00}.\label{eq:rlfm_zeroeq}
\end{aligned}
\end{equation}

Now we take the thermodynamic limit in which $M$, $N_f$, and $N_p$ tend towards infinity, but their ratios, $\alpha_f=N_f/M$ and $\alpha_p=N_p/M$, remain fixed.
We interpret the extra terms in Eq.~\eqref{eq:rlfm_plus0} as small perturbations to the auxiliary fields,
\begin{equation}
\delta \eta_J = \hat{u}_0 W_{0J}\qqc \delta \psi_j = \Delta y_0  X_{0 j}\qqc \delta \xi_a = \Delta \beta_0 X_{a 0}\qqc  \delta \zeta_j = - \hat{w}_0W_{j0},
\end{equation}
allowing us to expand each unknown quantity about its solution in the absence of the $0$-indexed variables (i.e., the solutions for $M$ data points, $N_f$ input features, and $N_p$ fit parameters),
\begin{equation}
\begin{aligned}
\hat{w}_J &\approx \hat{w}_{J\setminus 0} + \sum_K \nu^{\hat{w}}_{JK}\delta \eta_K + \sum_k\phi^{\hat{w}}_{Jk}\delta \psi_k + \sum_b \chi^{\hat{w}}_{Jb}\delta \xi_b + \sum_k\omega^{\hat{w}}_{Jk}\delta \zeta_k
\\
\hat{u}_j &\approx \hat{u}_{j\setminus 0} + \sum_K \nu^{\hat{u}}_{jK}\delta \eta_K + \sum_k\phi^{\hat{u}}_{jk}\delta \psi_k + \sum_b \chi^{\hat{u}}_{jb}\delta \xi_b + \sum_k\omega^{\hat{u}}_{jk}\delta \zeta_k
\\
\Delta y_a &\approx \Delta y_{a\setminus 0} + \sum_K \nu^{\Delta y}_{aK}\delta \eta_K + \sum_k\phi^{\Delta y}_{ak}\delta \psi_k + \sum_b \chi^{\Delta y}_{ab}\delta \xi_b + \sum_k\omega^{\Delta y}_{ak}\delta \zeta_k
\\
\Delta \beta_j&\approx\Delta \beta_{j\setminus 0} + \sum_K \nu^{\Delta \beta}_{jK}\delta \eta_K + \sum_k\phi^{\Delta \beta}_{jk}\delta \psi_k + \sum_b \chi^{\Delta \beta}_{jb}\delta \xi_b + \sum_k\omega^{\Delta \beta}_{jk}\delta \zeta_k.\label{eq:rlfm_cavexp}
\end{aligned}
\end{equation}
We define each of the susceptibility matrices as a derivative of a variable with respect to an auxiliary field,
\begin{equation}
\begin{alignedat}{4}
\nu^{\hat{w}}_{JK} &= \pdv{\hat{w}_J}{\eta_K}\qqc & \phi^{\hat{w}}_{Jk} &= \pdv{\hat{w}_J}{\psi_k}\qqc& \chi^{\hat{w}}_{Jb} &= \pdv{\hat{w}_J}{\xi_b}\qqc & \omega^{\hat{w}}_{Jk} &= \pdv{\hat{w}_J}{\zeta_k},\\
\nu^{\hat{u}}_{jK} &= \pdv{\hat{u}_j}{\eta_K}\qqc & \phi^{\hat{u}}_{jk} &= \pdv{\hat{u}_j}{\psi_k}\qqc& \chi^{\hat{u}}_{jb} &= \pdv{\hat{u}_j}{\xi_b}\qqc & \omega^{\hat{u}}_{jk} &= \pdv{\hat{u}_j}{\zeta_k},\\
\nu^{\Delta y}_{aK} &= \pdv{\Delta y_a}{\eta_K}\qqc  & \phi^{\Delta y}_{ak} &= \pdv{\Delta y_a}{\psi_k}\qqc & \chi^{\Delta y}_{ab} &= \pdv{\Delta y_a}{\xi_b}\qqc &\omega^{\Delta y}_{ak} &= \pdv{\Delta y_a}{\zeta_k},\\
\nu^{\Delta \beta}_{jK} &= \pdv{\Delta \beta_j}{\eta_K}\qqc & \phi^{\Delta\beta}_{jk} &= \pdv{\Delta \beta_j}{\psi_k}\qqc & \chi^{\Delta \beta}_{jb} &= \pdv{\Delta \beta_j}{\xi_b}\qqc & \omega^{\Delta\beta}_{jk} &= \pdv{\Delta\beta_j}{\zeta_k}.
\end{alignedat}
\end{equation}

\subsubsection{Central Limit Theorem} 

Substituting the expansions in Eq.~\eqref{eq:rlfm_cavexp} into the $0$-indexed equations in Eq.~\eqref{eq:rlfm_zeroeq}, we find that each of the resulting sums contains a thermodynamically large number of statistically uncorrelated terms. 
This means that each sum satisfies the conditions necessary to apply the central limit theorem,
allowing us to express each in terms of a single normally-distributed random variable described by just its mean and its variance.

First, we approximate each of the sums containing one of the unperturbed quantities, $\hat{w}_{J\setminus 0}$, $\hat{u}_{j\setminus 0}$, $ \Delta y_{a\setminus 0}$, or $\Delta \beta_{j\setminus 0}$.
The unperturbed quantities in each of these sums are statistically independent of all elements of both $X$ and $W$ with a $0$-valued index.
Using this fact, we find 
\begin{equation}
\begin{alignedat}{3}
\sum_k  \hat{u}_{k\setminus 0} W_{k0}  &\approx \sigma_{\hat{w}}z_{\hat{w}} \qqc& \sigma_{\hat{w}}^2 &= \sigma_W^2 \frac{\alpha_f}{\alpha_p}\expval*{\hat{u}^2} \qqc& \expval*{\hat{u}^2} &= \frac{1}{N_f}\sum_k\hat{u}_{k\setminus 0}^2\\
\sum_b \Delta y_{b\setminus 0} X_{b 0}  &\approx \sigma_{\hat{u} }z_{\hat{u}} \qqc&
\sigma_{\hat{u}}^2 &= \sigma_X^2\alpha_f^{-1}\expval*{\Delta y^2} \qqc& \expval*{\Delta y^2} &= \frac{1}{M}\sum_b  \Delta y_{b\setminus 0}^2\\
\sum_k \Delta \beta_{k\setminus 0} X_{0 k} &\approx \sigma_{\Delta y}z_{\Delta y} \qqc&
\sigma_{\Delta y}^2 &= \sigma_X^2 \expval*{\Delta \beta^2} \qqc & \expval*{\Delta\beta^2} &= \frac{1}{N_f}\sum_k \Delta \beta_{k\setminus 0}^2\\
\sum_K \hat{w}_{K\setminus 0}  W_{0 K} &\approx \sigma_{\Delta\beta}z_{\Delta\beta} \qqc&
\sigma_{\Delta\beta}^2 &= \sigma_W^2  \expval*{\hat{w}^2} \qqc& \expval*{\hat{w}^2}  &=\frac{1}{N_p}\sum_K \hat{w}_{K\setminus 0}^2,
\end{alignedat}
\end{equation}
where $z_{\hat{w}}$, $z_{\hat{u}}$,  $z_{\Delta y}$, and $z_{\Delta\beta}$ are all random variables with zero mean and unit variance and can easily be shown to be statistically independent from one another.

Note that we have used the same notation, $\expval*{\Delta y^2}$ and $\expval*{\Delta \beta^2}$, for the two averages defined previously in Sec.~\ref{sec:gen_sol} even though they each lack an ensemble average. In doing so, we have employed the ansatz that these sums will converge to their ensemble averages in the thermodynamic limit.
This assumption is typical of the cavity method.

Next, we approximate each of the sums containing one of the square susceptibility matrices.
Using the fact that all of the susceptibility matrices are statistically independent of all elements of both $X$ and $W$ with a $0$-valued index, we find that each of these sums is dominated by its mean with its variance going to zero in the thermodynamic limit,
\begin{equation}
\begin{alignedat}{2}
\sum_{jk}\omega^{\hat{u}}_{jk}W_{j0}W_{k0} &\approx \sigma_W^2 \frac{\alpha_f}{\alpha_p}\omega \qqc & \omega &= \frac{1}{N_f}\sum_k \omega^{\hat{u}}_{kk}\\
\sum_{ab}\chi^{\Delta y}_{ab}X_{a0}X_{b0} &\approx \sigma_X^2 \alpha_f^{-1}\chi \qqc & \chi &= \frac{1}{M}\sum_b\chi^{\Delta y}_{bb}\\
\sum_{jk}\phi^{\Delta \beta}_{jk}X_{0j}X_{0k} &\approx \sigma_X^2 \phi \qqc & \phi &= \frac{1}{N_f}\sum_k \phi^{\Delta \beta}_{kk}\\
\sum_{JK}\nu^{\hat{w}}_{JK}W_{0J}W_{0K} &\approx \sigma_W^2 \nu \qqc & \nu &=\frac{1}{N_p}\sum_K\nu^{\hat{w}}_{KK}
\end{alignedat}
\end{equation}
where $\omega$, $\chi$, $\phi$, and $\nu$ can be interpreted as a set of scalar susceptibilities.

Finally, it is straightforward to show that both the mean and variance each of the sums containing a rectangular susceptibility matrix goes to zero in the thermodynamic limit and can therefore be neglected.

\subsubsection{Self-consistency Equations}

Applying the approximations from the previous section,
we find a set of self-consistent equations for $\hat{w}_0$, $\hat{u}_0$, $\Delta y_0$, and $\Delta \beta_0$,
\begin{equation}
\begin{aligned}
\lambda \hat{w}_0 &\approx  \sigma_{\hat{w}}z_{\hat{w}} -\hat{w}_0\sigma_W^2 \frac{\alpha_f}{\alpha_p}\omega + \eta_0 \\
\hat{u}_0 &\approx \sigma_{ \hat{u} }z_{ \hat{u}} +  \Delta \beta_0\sigma_X^2 \alpha_f^{-1}\chi + \psi_0 \\
\Delta y_0 &\approx \sigma_{\Delta y}z_{\Delta y} + \Delta y_0\sigma_X^2 \phi + \delta y^*_{\mathrm{NL}}(\vbx_0) + \varepsilon_0 + \xi_0 \\
\Delta \beta_0 &\approx \beta_0 -  \sigma_{\Delta\beta}z_{\Delta\beta} - \hat{u}_0\sigma_W^2 \nu  +\zeta_0,
\end{aligned}
\end{equation}
where have also made use of the fact that the terms including $X_{00}$ or $W_{00}$ are infinitesimally small in the thermodynamic limit with zero mean and variances of $\order{1/N_f}$ and $\order{1/N_p}$, respectively.
Solving these equations for the $0$-indexed variables, we find
\begin{equation}
\begin{aligned}
\hat{w}_0  &= \frac{\sigma_{\hat{w}}z_{\hat{w}}  + \eta_0}{\lambda + \sigma_W^2 \frac{\alpha_f}{\alpha_p}\omega}\\
\hat{u}_0 &= \frac{\sigma_{\hat{u} }z_{\hat{u}} + \psi_0 + \sigma_X^2 \alpha_f^{-1}\chi\qty(\beta_0 - \sigma_{\Delta \beta}z_{\Delta \beta}   + \zeta_0)}{1 + \sigma_W^2\sigma_X^2\alpha_f^{-1}\chi\nu} \\
\Delta y_0 &= \frac{ \sigma_{\Delta y}z_{\Delta y}+ \delta y^*_{\mathrm{NL}}(\vbx_0) + \varepsilon_0 + \xi_0 }{1-\sigma_X^2 \phi} \\
\Delta \beta_0 &= \frac{\beta_0 - \sigma_{\Delta \beta}z_{\Delta \beta}   +\zeta_0 -\sigma_W^2\nu^2\qty(\sigma_{\hat{u} }z_{\hat{u}} + \psi_0)}{1 + \sigma_W^2\sigma_X^2\alpha_f^{-1}\chi\nu}.\label{eq:rlfm_zeroselfcon}
\end{aligned}
\end{equation}

Next, we derive a set of self-consistent equations for the scalar susceptibilities by taking appropriate derivatives of these variables with respect to the auxiliary fields,
\begin{equation}
\begin{aligned}
\nu &= \frac{1}{N_p}\sum_K\nu^{\hat{w}}_{KK} \approx \E\qty[\nu^{\hat{w}}_{00}] = \E\qty[\pdv{\hat{w}_0}{\eta_0}] = \frac{1}{\lambda + \sigma_W^2 \frac{\alpha_f}{\alpha_p}\omega}\\
\omega &= \frac{1}{N_f}\sum_k \omega^{\hat{u}}_{kk} \approx \E\qty[\omega^{\hat{u}}_{00}] = \E\qty[\pdv{\hat{u}_0}{\zeta_0}] = \frac{\sigma_X^2\alpha_f^{-1}\chi}{1+\sigma_W^2\sigma_X^2\alpha_f^{-1}\chi\nu}\\
\chi &=\frac{1}{M}\sum_b\chi^{\Delta y}_{bb}\approx  \E\qty[\chi^{\Delta y}_{00}] = \E\qty[\pdv{\Delta y_0}{\xi_0}] = \frac{1}{1-\sigma_X^2\phi}\\
\phi &= \frac{1}{N_f}\sum_k \phi^{\Delta \beta}_{kk} \approx  \E\qty[\phi^{\Delta \beta}_{00}] = \E\qty[\pdv{\Delta\beta_0}{\psi_0}] = -\frac{\sigma_W^2\nu}{1+\sigma_W^2\sigma_X^2\alpha_f^{-1}\chi\nu}.
\end{aligned}
\end{equation}
Furthermore, we note that there are two additional derivatives that have not yet appeared in the calculation up to this point, $\partial\hat{u}_j/\partial\psi_j$ and $\partial\Delta\beta_j/\partial\zeta_j$.
It is clear to see from the equations for $\hat{u}_0$ and $\Delta\beta_0$ that these two additional derivatives are equivalent.
Evaluating these derivatives, we define a fifth scalar susceptibility,
\begin{equation}
\kappa = \frac{1}{N_f}\sum_k \phi^{\hat{u}}_{kk} = \frac{1}{N_f}\sum_k \omega^{\Delta\beta}_{kk} \approx \E\qty[\pdv{\hat{u}_0}{\psi_0}]  = \E\qty[\pdv{\Delta\beta_0}{\zeta_0}] = \frac{1}{1+\sigma_W^2\sigma_X^2\alpha_f^{-1}\chi\nu}.\label{eq:rlfm_susceptselfconkappa}
\end{equation}
Using this formula for $\kappa$, we re-express the four other susceptibilities as
\begin{equation}
\omega = \sigma_X^2\alpha_f^{-1}\chi\kappa\qqc
\phi =  -\sigma_W^2\nu\kappa\qqc\chi = \frac{1}{1+\sigma_W^2\sigma_X^2\nu\kappa}\qqc
\nu = \frac{1}{\lambda + \sigma_W^2\sigma_X^2  \alpha_p^{-1}\chi\kappa}.\label{eq:rlfm_susceptselfcon}
\end{equation}
Finally, we square and average each of the expressions in Eq.~\eqref{eq:rlfm_zeroselfcon} to find self-consistent equations for the four mean squared averages (setting the auxiliary fields to zero),
\begin{equation}
\begin{aligned}
\expval*{\hat{w}^2} &= \frac{1}{N_p}\sum_K \hat{w}_{K\setminus 0}^2 \approx \E\qty[\hat{w}_0^2] = \nu^2\sigma_W^2 \frac{\alpha_f}{\alpha_p}\expval*{\hat{u}^2} \\
\expval*{\hat{u}^2} &= \frac{1}{N_f}\sum_k \hat{u}_{k\setminus 0}^2 \approx\E\qty[\hat{u}_0^2] =  \kappa^2\sigma_X^2\alpha_f^{-1}\expval*{\Delta y^2}  + \omega^2\qty(\sigma_\beta^2 + \sigma_W^2\expval*{\hat{w}^2})\\
\expval*{\Delta y^2} &= \frac{1}{M}\sum_b \Delta y_{b\setminus 0}^2 \approx\E\qty[\Delta y_0^2] = \chi^2 \qty(\sigma_X^2\expval*{\Delta\beta^2} + \sigma_{\delta y^*}^2 + \sigma_\varepsilon^2) \\
\expval*{\Delta\beta^2} &=\frac{1}{N_f}\sum_k \Delta \beta_{k\setminus 0}^2 \approx \E\qty[\Delta \beta_0^2] = \kappa^2\qty( \sigma_\beta^2 + \sigma_W^2\expval*{\hat{w}^2}) + \phi^2\sigma_X^2\alpha_f^{-1} \expval*{\Delta y^2} .\label{eq:rlfm_squareselfcon}
\end{aligned}
\end{equation}

\subsubsection{Solution with Finite Regularization ($\lambda \sim 1$)} 

Next, we derive the solutions when the regularization parameter $\lambda$ is finite.
To do this, we combine the self-consistency equations for the susceptibilities in Eqs.~\eqref{eq:rlfm_susceptselfconkappa} and \eqref{eq:rlfm_susceptselfcon} to derive a cubic equation for $\chi$,
\begin{equation}
0 =  \chi^3 +  (\alpha_f + \alpha_p - 2)\chi^2 + \qty[ \qty(\alpha_f-1)\qty(\alpha_p-1) +\alpha_f\alpha_p \bar{\lambda} ]\chi  -\alpha_f\alpha_p\bar{\lambda},\label{eq:rlfm_poly} 
\end{equation}
where we have defined the dimensionless regularization parameter
\begin{equation}
\bar{\lambda} = \frac{\lambda}{\sigma_W^2\sigma_X^2}.
\end{equation}
This cubic equation indicates that we should expect three different solutions for $\chi$.
Using these solutions, we can derive the associated solutions for the rest of the susceptibilities.
Furthermore, we solve Eq.~\eqref{eq:rlfm_squareselfcon} to find
\begin{equation}
\mqty(\expval*{\hat{w}^2} \\ \expval*{\hat{u}^2} \\ \expval*{\Delta y^2}\\ \expval*{\Delta\beta^2}) = 
\mqty(1 & -\sigma_W^2 \frac{\alpha_f}{\alpha_p}\nu^2 & 0 & 0\\
-\sigma_W^2\omega^2 & 1 & -\sigma_X^2\alpha_f^{-1}\kappa^2 & 0\\
0 & 0 & 1 & -\sigma_X^2\chi^2\\
-\sigma_W^2\kappa^2 & 0 &  -\sigma_X^2 \alpha_f^{-1} \phi^2 & 1)^{-1}
\mqty(0 \\ \sigma_\beta^2\omega^2 \\ (\sigma_\varepsilon^2 + \sigma_{\delta y^*}^2)\chi^2 \\ \sigma_\beta^2\kappa^2 ).
\end{equation}
In combination with the solutions for the five scalar susceptibilities, these solutions are exact in the thermodynamic limit.

\subsubsection{Solutions in Ridge-less Limit ($\lambda\rightarrow 0$)}

Next, we take the ridge-less limit in which $\lambda \rightarrow 0$.  
Based on the cubic equation for $\chi$ in  Eq.~\eqref{eq:rlfm_poly}, we make the ansatz that the lowest order contribution to $\chi$ is $\order{1}$ in small $\bar{\lambda}$,
\begin{equation}
\chi \approx \chi_0 + \bar{\lambda} \chi_1.
\end{equation}
We then expand Eq.~\eqref{eq:rlfm_poly} in orders of $\lambda$ to find solutions for $\chi_0$ and $\chi_1$. 
Using these solutions, we solve for the following coefficients for the remaining susceptibilities.
\begin{equation}
\begin{aligned}
\nu &\approx \frac{1}{\bar{\lambda}}\nu_{-1} + \nu_0\\
\kappa &\approx \kappa_0 + \bar{\lambda}\kappa_1\\
\phi &\approx \frac{1}{\bar{\lambda}}\phi_{-1} + \phi_0\\
\omega &\approx \omega_0 + \bar{\lambda}\omega_1,
\end{aligned}
\end{equation}
Finally, we expand the mean squared averages in small $\lambda$ as
\begin{equation}
\begin{aligned}
\expval*{\hat{w}^2} &\approx \expval*{\hat{w}^2}_0 + \bar{\lambda}^2\expval*{\hat{w}^2}_2\\
\expval*{\hat{u}^2} &\approx \expval*{\hat{u}^2}_0 + \bar{\lambda}^2\expval*{\hat{u}^2}_2\\
\expval*{\Delta y^2} &\approx \expval*{\Delta y^2}_0 + \bar{\lambda}^2 \expval*{\Delta y^2}_2\\
\expval*{\Delta \beta^2} &\approx \expval*{\Delta \beta^2}_0 + \bar{\lambda}^2 \expval*{\Delta \beta^2}_2,
\end{aligned}
\end{equation}
and then use the solutions for the susceptibilities to solve Eq.~\eqref{eq:rlfm_squareselfcon} for these coefficients.

We find three sets of solutions for all quantities, corresponding to the three regimes of the random linear features model.
To determine when each solution applies,  we use the fact that each of the ensemble-averaged quantities $\expval*{\hat{w}^2}$,  $\expval*{\hat{u}^2} $, $\expval*{\Delta y^2}$, and $\expval*{\Delta \beta^2}$ must be positive. All together, we find the solutions for the  ensemble-averaged squared quantities in the $\lambda\rightarrow 0$ limit to be
\begingroup
\allowdisplaybreaks
\begin{align}
\expval*{\hat{w}^2} &=\left\{\begin{array}{cl}
\frac{\sigma_\beta^2}{\sigma_W^2} \frac{\alpha_f}{(\alpha_p - \alpha_f)} + \frac{(\sigma_\varepsilon^2 + \sigma_{\delta y^*}^2)}{\sigma_W^2\sigma_X^2} \frac{\alpha_f^2}{(1-\alpha_f)(\alpha_p-\alpha_f)} & \qif N_f < N_p, M\\
\frac{\sigma_\beta^2}{\sigma_W^2} \frac{(1-\alpha_p + \alpha_f-\alpha_p)}{(1-\alpha_p)(\alpha_f - \alpha_p)} + \frac{(\sigma_\varepsilon^2 + \sigma_{\delta y^*}^2)}{\sigma_W^2\sigma_X^2} \frac{\alpha_f\alpha_p}{(1-\alpha_p)(\alpha_f-\alpha_p)}  & \qif N_p < N_f, M\\
 \frac{\sigma_\beta^2}{\sigma_W^2} \frac{1}{(\alpha_p-1)} + \frac{(\sigma_\varepsilon^2 + \sigma_{\delta y^*}^2)}{\sigma_W^2\sigma_X^2} \frac{\alpha_f}{(\alpha_f-1)(\alpha_p-1)}   & \qif M < N_f, N_p
\end{array} \right. \\
\expval*{\hat{u}^2} &= \left\{\begin{array}{cl}
\frac{\lambda^2}{\sigma_X^4\sigma_W^4}\qty[\sigma_\beta^2\sigma_X^4 \frac{\alpha_p^3}{(\alpha_p-\alpha_f)^3} +\sigma_X^2  (\sigma_\varepsilon^2 + \sigma_{\delta y^*}^2)\frac{\alpha_f\alpha_p^3}{(1-\alpha_f)(\alpha_p-\alpha_f)^3}] & \qif N_f < N_p, M\\
\sigma_\beta^2\sigma_X^4 \frac{(1-\alpha_p)(\alpha_f-\alpha_p)(1-\alpha_p + \alpha_f-\alpha_p)}{\alpha_f^3} + \sigma_X^2 (\sigma_\varepsilon^2 + \sigma_{\delta y^*}^2) \frac{(1-\alpha_p)(\alpha_f-\alpha_p)}{\alpha_f^2} & \qif N_p < N_f, M\\
\frac{\lambda^2}{\sigma_X^4\sigma_W^4} \qty[\sigma_\beta^2\sigma_X^4 \frac{\alpha_p^3}{\alpha_f(\alpha_p-1)^3} + \sigma_X^2 (\sigma_\varepsilon^2 + \sigma_{\delta y^*}^2)  \frac{\alpha_p^3}{(\alpha_f-1)(\alpha_p-1)^3}]   & \qif M < N_f, N_p
\end{array} \right. \\
\expval*{\Delta y^2} &=  \left\{
\begin{array}{cl}
(\sigma_\varepsilon^2 + \sigma_{\delta y^*}^2)  (1-\alpha_f)& \qif N_f < N_p, M\\
\sigma_\beta^2\sigma_X^2 \frac{(1-\alpha_p)(\alpha_f-\alpha_p)}{\alpha_f}  + (\sigma_\varepsilon^2 + \sigma_{\delta y^*}^2) (1-\alpha_p) & \qif N_p < N_f, M\\
\frac{\lambda^2}{\sigma_X^4\sigma_W^4}\qty[\sigma_\beta^2\sigma_X^2\frac{\alpha_f\alpha_p^3}{(\alpha_f-1)(\alpha_p-1)^3}  + (\sigma_\varepsilon^2 + \sigma_{\delta y^*}^2)  \frac{\alpha_f^2\alpha_p^2(\alpha_f-1+\alpha_p-1)}{(\alpha_f-1)^3(\alpha_p-1)^3}] & \qif M < N_f, N_p
\end{array}
\right.\\
\expval*{\Delta\beta^2} &=  \left\{
\begin{array}{cl}
\frac{(\sigma_\varepsilon^2 + \sigma_{\delta y^*}^2) }{\sigma_X^2} \frac{\alpha_f}{(1-\alpha_f)} & \qif N_f < N_p, M\\
\sigma_\beta^2\frac{(\alpha_f-\alpha_p)}{\alpha_f(1-\alpha_p)} +  \frac{(\sigma_\varepsilon^2 + \sigma_{\delta y^*}^2) }{\sigma_X^2} \frac{\alpha_p}{(1-\alpha_p)} & \qif N_p < N_f, M\\
 \sigma_\beta^2\frac{\alpha_p(\alpha_f-1)}{\alpha_f(\alpha_p-1)} +  \frac{(\sigma_\varepsilon^2 + \sigma_{\delta y^*}^2) }{\sigma_X^2} \frac{(\alpha_f-1+\alpha_p-1)}{(\alpha_f-1)(\alpha_p-1)} & \qif M < N_f, N_p.
\end{array}
\right.
\end{align}
In addition, to lowest order in small $\lambda$, the five scalar susceptibilities are
\begin{align}
\chi &= \left\{
\begin{array}{cl}
1-\alpha_f & \qif N_f < N_p, M\\
1-\alpha_p & \qif N_p < N_f, M\\
\frac{\lambda}{\sigma_W^2\sigma_X^2} \frac{\alpha_f\alpha_p}{(1-\alpha_f)(1-\alpha_p)} & \qif M < N_f, N_p
\end{array}
\right.\\
\nu &= \left\{
\begin{array}{cl}
\frac{1}{\lambda}\frac{(\alpha_p-\alpha_f)}{\alpha_p} + \frac{1}{\sigma_W^2\sigma_X^2}\frac{\alpha_f^2}{ \qty(1-\alpha_f)\qty(\alpha_p-\alpha_f)} & \qif N_f < N_p, M\\
\frac{1}{\sigma_W^2\sigma_X^2} \frac{\alpha_f\alpha_p}{(1-\alpha_p)(\alpha_f-\alpha_p)} & \qif N_p < N_f, M\\
\frac{1}{\lambda}\frac{(\alpha_p-1)}{\alpha_p} + \frac{1}{\sigma_W^2\sigma_X^2} \frac{\alpha_f}{ \qty(\alpha_f-1)\qty(\alpha_p-1)}& \qif M < N_f, N_p
\end{array}
\right.\\
\kappa &= \left\{
\begin{array}{cl}
\frac{\lambda}{\sigma_W^2\sigma_X^2} \frac{\alpha_f\alpha_p}{(1-\alpha_f)(\alpha_p-\alpha_f)} & \qif N_f < N_p, M\\
\frac{(\alpha_f-\alpha_p)}{\alpha_f} & \qif N_p < N_f, M\\
\frac{(\alpha_f-1)}{\alpha_f} & \qif M < N_f, N_p
\end{array}
\right.\\
\omega &= \left\{
\begin{array}{cl}
\frac{\lambda}{\sigma_W^2} \frac{\alpha_p}{(\alpha_p-\alpha_f)} & \qif N_f < N_p, M\\
\sigma_X^2\frac{(1-\alpha_p)(\alpha_f-\alpha_p)}{\alpha_f^2} & \qif N_p < N_f, M\\
\frac{\lambda}{\sigma_W^2} \frac{\alpha_p}{\alpha_f(\alpha_p-1)} & \qif M < N_f, N_p
\end{array}
\right.\\
\phi &= \left\{
\begin{array}{cl}
-\frac{1}{\sigma_X^2}\frac{\alpha_f}{ (1-\alpha_f)} & \qif N_f < N_p, M\\
-\frac{1}{\sigma_X^2}\frac{\alpha_p}{ (1-\alpha_p)} & \qif N_p < N_f, M\\
-\frac{\sigma_W^2}{\lambda}\frac{(\alpha_f-1)(\alpha_p-1)}{\alpha_f \alpha_p} & \qif M < N_f, N_p.
\end{array}
\right.
\end{align}
\endgroup

We use the quantities $\expval*{\Delta y^2}$ and $\expval*{\Delta \beta^2}$ above in combination with formulas for the training and test error in Sec.~\ref{sec:gen_sol} to obtain the expressions in Eqs.~\eqref{eq:rlfm_train} and \eqref{eq:rlfm_test}.

\subsubsection{Bias-Variance Decomposition}\label{sec:rlfm_bv}

Next, we derive the bias and variance. 
According to the general solutions in Eqs.~\eqref{eq:SIgen_bias} and \eqref{eq:SIgen_var}, we require the quantity $\expval*{\Delta\beta_1\Delta\beta_2}$.
To calculate $\expval*{\Delta\beta_1\Delta\beta_2}$, we apply the self-consistent equations, Eq.~\eqref{eq:rlfm_zeroselfcon}, to two models each trained separately on one of two independent training sets, with all other random variables held in common.
We specify which quantities depend on each of the training sets using a subscript $1$ or $2$ for data sets $\mathcal{D}_1$ and $\mathcal{D}_2$, respectively.
For training set $\mathcal{D}_1$, these equations are
\begin{equation}
\begin{aligned}
\hat{w}_{1,0}  &= \nu\sigma_{\hat{w}}z_{\hat{w}_1}\\
\hat{u}_{1,0} &= \kappa \sigma_{\hat{u}}z_{\hat{u}_1} + \omega\qty(\beta_0 - \sigma_{\Delta \beta}z_{\Delta \beta_1} )\\
\Delta y_{1,0} &= \chi\qty(\sigma_{\Delta y}z_{\Delta y_1} + \delta y^*_{\mathrm{NL}}(\vbx_{1, 0}) + \varepsilon_{1,0})\\
\Delta \beta_{1,0} &= \kappa \qty(\beta_0 - \sigma_{\Delta \beta}z_{\Delta \beta_1}) + \phi  \sigma_{\hat{u}}z_{\hat{u}_1},
\end{aligned}
\end{equation}
while for training set $\mathcal{D}_2$, they are
\begin{equation}
\begin{aligned}
\hat{w}_{2,0}  &= \nu\sigma_{\hat{w}}z_{\hat{w}_2}\\
\hat{u}_{2,0} &= \kappa \sigma_{\hat{u}}z_{\hat{u}_2} + \omega\qty(\beta_0 - \sigma_{\Delta \beta}z_{\Delta \beta_2} )\\
\Delta y_{2,0} &= \chi\qty(\sigma_{\Delta y}z_{\Delta y_2} + \delta y^*_{\mathrm{NL}}(\vbx_{2, 0}) + \varepsilon_{2,0})\\
\Delta \beta_{2,0} &= \kappa \qty(\beta_0 - \sigma_{\Delta \beta}z_{\Delta \beta_2}) + \phi \sigma_{\hat{u}}z_{\hat{u}_2}.
\end{aligned}
\end{equation}
Multiplying these equations and making the self-averaging approximation, we find equations for the covariance of each of the unknown variables,
\begin{equation}
\begin{aligned}
\expval*{\hat{w}_1\hat{w}_2} &= \frac{1}{N_p}\sum_K \hat{w}_{1, K}\hat{w}_{2, K} \approx \E\qty[\hat{w}_{1, 0}\hat{w}_{2, 0}] = \nu^2\E\qty[\sigma_{\hat{w}}^2z_{\hat{w}_1}z_{\hat{w}_2}] \\
\expval*{\hat{u}_1\hat{u}_2} &= \frac{1}{N_f}\sum_k \hat{u}_{1, k}\hat{u}_{2, k} \approx \E\qty[\hat{u}_{1, 0}\hat{u}_{2, 0}] = \kappa^2\E\qty[\sigma_{\hat{u}}^2z_{\hat{u}_1}z_{\hat{u}_2}] + \omega^2\qty(\sigma_\beta^2 + \E\qty[\sigma_{\Delta \beta}^2z_{\Delta \beta_1}z_{\Delta \beta_2}] ) \\
\expval*{\Delta y_1 \Delta y_2} &= \frac{1}{M}\sum_b \Delta y_{1, b}\Delta y_{2, b} \approx \E\qty[\Delta y_{1, 0}\Delta y_{2, 0}] = \chi^2\E\qty[\sigma_{\Delta y}^2z_{\Delta y_1}z_{\Delta y_2}] \\
\expval*{\Delta\beta_1\Delta\beta_2} &= \frac{1}{N_f}\sum_k \Delta \beta_{1, k}  \Delta \beta_{2, k} \approx \E\qty[ \Delta \beta_{1, 0}  \Delta \beta_{2, 0}] = \kappa^2\qty(\sigma_\beta^2 + \E\qty[\sigma_{\Delta \beta}^2z_{\Delta \beta_1}z_{\Delta \beta_2}]) + \phi^2\E\qty[\sigma_{\hat{u}}^2z_{\hat{u}_1}z_{\hat{u}_2}] .\label{eq:rlfm_corrselfcon}
\end{aligned}
\end{equation}
Next, we calculate each of the four resulting expectation values of products of random variables.
Converting each of the random variables $z_{\hat{w}_1}$, $z_{\Delta \beta_1}$, etc.,  back into their forms as sums, we use the independence of elements of the random matrices and the other variables to find
\begin{equation}
\begin{aligned}
\E\qty[\sigma_{\hat{w}}^2z_{\hat{w}_1}z_{\hat{w}_2}] &\approx \E\qty[\sum_{jk}\hat{u}_{1, j\setminus 0}\hat{u}_{2, k\setminus 0}W_{j0}W_{k0}]= \sigma_W^2 \frac{\alpha_f}{\alpha_p} \expval*{\hat{u}_1\hat{u}_2}\\
\E\qty[\sigma_{\Delta \beta}^2z_{\Delta \beta_1}z_{\Delta \beta_2}]&\approx \E\qty[\sum_{JK}\hat{w}_{1, J\setminus 0}\hat{w}_{2, K\setminus 0}W_{0J}W_{0K}] =\sigma_W^2 \expval*{\hat{w}_1\hat{w}_2}\\
\E\qty[\sigma_{\hat{u}}^2z_{\hat{u}_1}z_{\hat{u}_2}] &\approx \E\qty[\sum_{ab}\Delta y_{1, a\setminus 0}\Delta y_{2, b\setminus 0}X_{1, a0}X_{2, b0}] = 0\\
\E\qty[\sigma_{\Delta y}^2z_{\Delta y_1}z_{\Delta y_2}] &\approx \E\qty[\sum_{jk}\Delta\beta_{1, j\setminus 0}\Delta\beta_{2, k\setminus 0}X_{1, 0j}X_{2, 0k}]= 0.
\end{aligned}
\end{equation}
Substituting these results back into Eq.~\eqref{eq:rlfm_corrselfcon}, we find the self-consistent equations
\begin{equation}
\begin{aligned}
\expval*{\hat{w}_1\hat{w}_2} &= \nu^2\sigma_W^2\frac{\alpha_f}{\alpha_p} \expval*{\hat{u}_1\hat{u}_2}\\
\expval*{\hat{u}_1\hat{u}_2} &= \omega^2\qty(\sigma_\beta^2 + \sigma_W^2 \expval*{\hat{w}_1\hat{w}_2})\\
\expval*{\Delta y_1 \Delta y_2} &= 0\\
\expval*{\Delta\beta_1\Delta\beta_2} &= \kappa^2\qty( \sigma_\beta^2 + \sigma_W^2 \expval*{\hat{w}_1\hat{w}_2}).
\end{aligned}
\end{equation}
Next, we make the ansatz that the ensemble-averaged covariances are $\order{1}$ in small $\bar{\lambda}$ with the next order terms at $\order{\bar{\lambda}^2}$, 
\begin{equation}
\begin{aligned}
\expval*{\hat{w}_1\hat{w}_2} &\approx \expval*{\hat{w}_1\hat{w}_2}_0 + \bar{\lambda}^2\expval*{\hat{w}_1\hat{w}_2}_2\\
\expval*{\hat{u}_1\hat{u}_2}&\approx \expval*{\hat{u}_1\hat{u}_2}_0 + \bar{\lambda}^2 \expval*{\hat{u}_1\hat{u}_2}_2\\
\expval*{\Delta\beta_1\Delta\beta_2} &\approx \expval*{\Delta\beta_1\Delta\beta_2}_0 + \bar{\lambda}^2\expval*{\Delta\beta_1\Delta\beta_2}_2.
\end{aligned}
\end{equation}
All together, the covariances in the limit $\lambda\rightarrow 0$ are
\begingroup
\allowdisplaybreaks
\begin{align}
\expval*{\hat{w}_1\hat{w}_2} &=\left\{\begin{array}{cl}
 \frac{\sigma_\beta^2}{\sigma_W^2}\frac{\alpha_f}{(\alpha_p-\alpha_f)} & \qif N_f < N_p, M\\
 \frac{\sigma_\beta^2}{\sigma_W^2}\frac{\alpha_p}{(\alpha_f-\alpha_p)}  & \qif N_p < N_f, M\\
  \frac{\sigma_\beta^2}{\sigma_W^2}\frac{1}{(\alpha_f\alpha_p-1)}   & \qif M < N_f, N_p
\end{array} \right. \\
\expval*{\hat{u}_1\hat{u}_2} &= \left\{\begin{array}{cl}
\frac{\lambda^2}{\sigma_X^4\sigma_W^4}\sigma_\beta^2\sigma_X^4 \frac{\alpha_p^3}{(\alpha_p-\alpha_f)^3} & \qif N_f < N_p, M\\
 \sigma_\beta^2\sigma_X^4 \frac{(1-\alpha_p)^2(\alpha_f-\alpha_p)}{\alpha_f^3} & \qif N_p < N_f, M\\
\frac{\lambda^2}{\sigma_X^4\sigma_W^4}  \sigma_\beta^2\sigma_X^4 \frac{\alpha_p^3}{\alpha_f(\alpha_p-1)^2(\alpha_f\alpha_p-1)}  & \qif M < N_f, N_p
\end{array} \right. \\
\expval*{\Delta y_1\Delta y_2} &=  0\\
\expval*{\Delta\beta_1\Delta\beta_2} &=  \left\{
\begin{array}{cl}
\frac{\lambda^2}{\sigma_X^4\sigma_W^4} \sigma_\beta^2 \frac{\alpha_f^2\alpha_p^3}{(1-\alpha_f)^2(\alpha_p-\alpha_f)^3} & \qif N_f < N_p, M\\
\sigma_\beta^2 \frac{(\alpha_f-\alpha_p)}{\alpha_f} & \qif N_p < N_f, M\\
\sigma_\beta^2 \frac{\alpha_p(\alpha_f-1)^2}{\alpha_f(\alpha_f\alpha_p-1)} & \qif M < N_f, N_p.
\end{array}
\right.
\end{align}
\endgroup
Finally, we use the solution for $\expval*{\Delta\beta_1\Delta\beta_2}$ to find the bias and variance according to Eqs.~\eqref{eq:SIgen_bias} and \eqref{eq:SIgen_var}, 
resulting in Eqs.~\eqref{eq:rlfm_bias} and \eqref{eq:rlfm_var}.

\section{Spectral Densities of Kernel Matrices}

Here, we derive the spectral densities for the kernel matrix $Z^TZ$ using the technique laid out in Ref.~\onlinecite{Cui2020}.
According to this formalism, the spectral density of the kernel can be written in terms of the scalar susceptibility $\nu$, defined in the previous section, using the formula
\begin{align}
\rho(x) &= -\frac{1}{\pi}\lim_{\varepsilon\rightarrow 0^+}\Im \nu(-x + i\varepsilon).\label{eq:spectrum}
\end{align}

In addition, we expect there to be delta function of eigenvalues located at zero.
Although the above formula can in principle be used to obtain the fraction of eigenvalues at zero,
for convenience, we instead use the scalar susceptibility $\chi$, which can be shown to be exactly
\begin{equation}
\chi =\frac{1}{M}\Tr \pdv{\Delta \vby}{\vbxi} = \frac{1}{M}\Tr \qty[I_M - ZZ^+] = 1 - \frac{1}{M}\rank(Z^TZ).
\end{equation}
The fraction of eigenvalues at zero is then
\begin{align}
f_{\mathrm{zero}} = 1-\frac{1}{N_p}\rank(Z^TZ) =  \frac{\chi + \alpha_p -1}{\alpha_p}.\label{eq:fzero}
\end{align}

Next, we define dimensional versions of $\nu$ and $\lambda$,
\begin{equation}
\bar{\nu} = \sigma_W^2\sigma_X^2\nu\qqc \bar{\lambda} = \frac{\lambda}{\sigma_W^2\sigma_X^2}.
\end{equation}
Using the self-consistent equations for the scalar susceptibilities in Eqs.~\eqref{eq:rlfm_susceptselfconkappa} and \eqref{eq:rlfm_susceptselfcon}, we find a cubic equation for $\bar{\nu}$,
\begin{equation}
0  = (\alpha_p\bar{\lambda}\bar{\nu})^3 + \qty[1-\alpha_p + \alpha_f - \alpha_p](\alpha_p\bar{\lambda}\bar{\nu})^2 + \qty[(1-\alpha_p)(\alpha_f-\alpha_p) + \alpha_f\alpha_p\bar{\lambda}](\alpha_p\bar{\lambda}\bar{\nu}) - \alpha_f\alpha_p^2\bar{\lambda}.
\end{equation}
Solving this cubic equation analytically is very involved, so we refer to the solution in Ref.~\onlinecite{Dupic2014}.
Instead, we solve this equation numerically for the negative imaginary roots of $\nu(\lambda)$ with $\lambda = -x$, according to Eq.~\eqref{eq:spectrum}.
However, we also need to find the interval over which the eigenvalue spectrum is positive.
To do this, we rewrite the equation in general form for $\alpha_p\bar{\lambda}\bar{\nu}$,
\begin{equation}
(\alpha_p\bar{\lambda}\bar{\nu})^3 + a_2(\alpha_p\bar{\lambda}\bar{\nu})^2 + a_1(\alpha_p\bar{\lambda}\bar{\nu}) + a_0=0,
\end{equation}
where the coefficients are
\begin{equation}
\begin{aligned}
a_0 &= - \alpha_f\alpha_p^2\bar{\lambda}\\
a_1 &= (1-\alpha_p)(\alpha_f-\alpha_p) + \alpha_f\alpha_p\bar{\lambda}\\
a_2 &=  1-\alpha_p + \alpha_f - \alpha_p.
\end{aligned}
\end{equation}
The discriminant for a cubic equation is expressed in terms of these coefficients as
\begin{equation}
D(\lambda) = R^2 - Q^3
\end{equation}
with
\begin{equation}
\begin{aligned}
Q &= \frac{1}{9}\qty(a_2^2 - 3a_1)\\
R &= \frac{1}{54}\qty(9a_2a_1 - 27 a_0 - 2a_2^3).
\end{aligned}
\end{equation}
To find the limiting eigenvalues, we then solve the equation $D(\lambda) = 0$ (with $\lambda = -x$) numerically for the largest and smallest non-negative real roots.

To find the weight of the delta function component at zero, we use the solution for $\chi$ that we found previously, giving us
\begin{equation}
f_{\mathrm{zero}} = \left\{
\begin{array}{cl}
1- \frac{\alpha_f}{\alpha_p} & \qif N_f < N_p, M\\
0 & \qif N_p < N_f, M\\
1 - \alpha_p^{-1} & \qif M < N_f, N_p
\end{array}
\right.
= \max\qty(0, 1- \frac{\alpha_f}{\alpha_p}, 1 - \alpha_p^{-1} ).
\end{equation}

\section{Numerical Simulation Details}

\begin{figure}[h!]
\centering
\includegraphics[width=\linewidth]{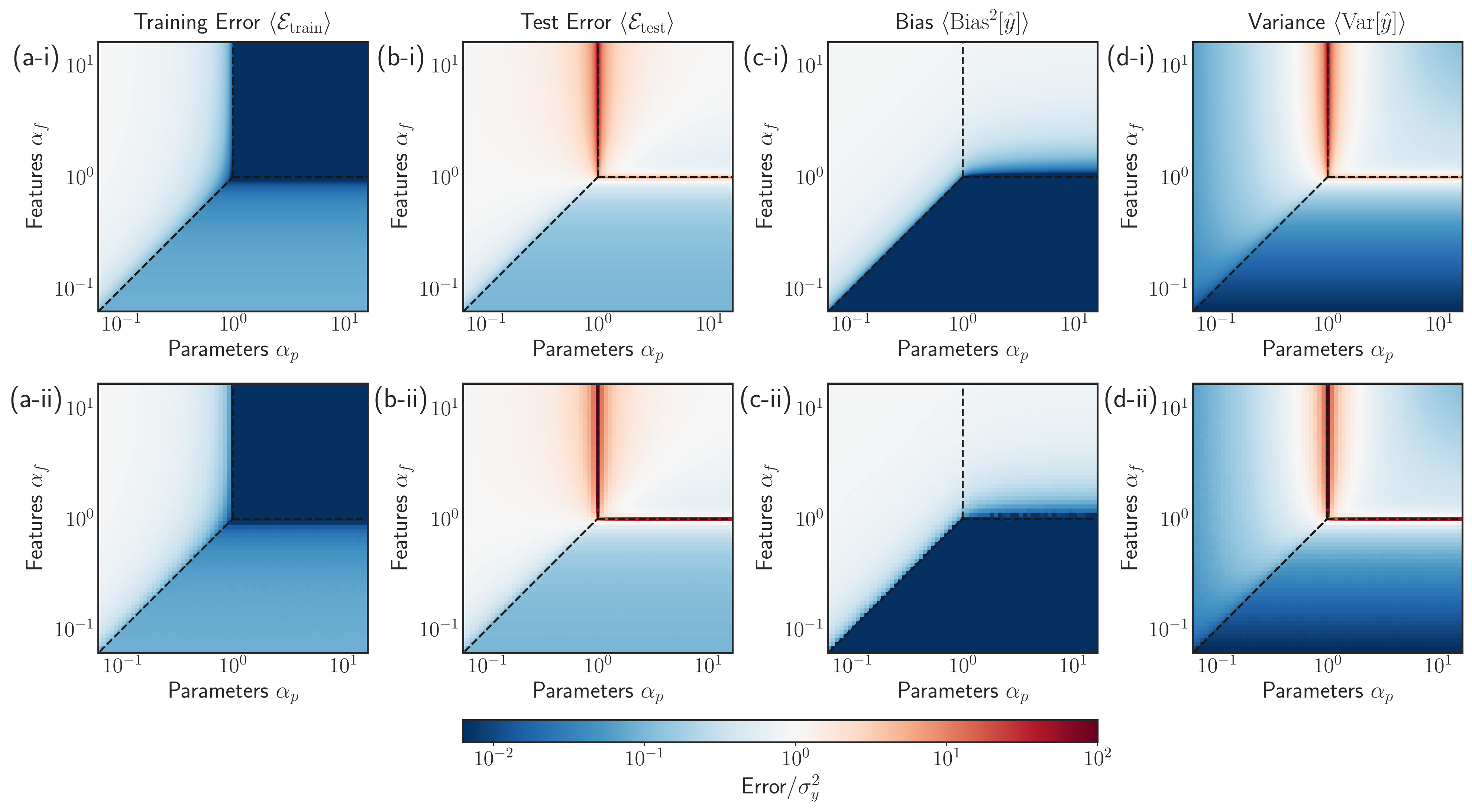} 
\caption{{\bf Comparison of analytic and numerical results for the random linear features model: Training error and bias-variance decomposition.}
(Top Row) Analytic solutions and (Bottom Row) numerical results are shown as a function of $\alpha_p=N_p/M$ and $\alpha_f=N_f/M$. 
Plotted are the ensemble-averaged (a) training error, (b) test error, (c) squared bias, and (d) variance. 
In each panel, black dashed lines show boundaries between different regimes of solutions depending on which is smallest of the quantities $M$, $N_f$, or $N_p$. 
The vertical and horizontal lines bound the interpolation, or overparameterized, regime, located at $\alpha_p > 1$ and $\alpha_f > 1$, while the diagonal line marks the boundary between the large bias and minimal bias underparameterized regimes for a linear teacher model.
All solutions have been scaled by the variance of the training set labels $\sigma_y^2 = \sigma_\beta^2\sigma_X^2 +\sigma_\varepsilon^2$.
}\label{fig:rlfm_numerics1}
\end{figure}

In this section, we explain our procedures for generating numerical results.
Fig.~\ref{fig:rlfm_numerics1} provides comparisons to numerical results for the training error, test error, bias, and variance.

\subsection{General Details}\label{sec:numdetails}

In all plots of training error, test error, bias, and variance, each point (or pixel for 2$d$ plots)  is averaged over $1000$ independent simulations, unless located exactly at a phase transition, in which case, each point is averaged over $150000$ simulations.
Small error bars are shown each plot, representing the error on the mean.
We also scale the error in each plot by the variance of the labels $\sigma_y^2 = \sigma_\beta^2\sigma_X^2 + \sigma_{\delta y^*}^2 +\sigma_\varepsilon^2$.
In all simulations, we use training and test sets of size $M=M'=512$,  a signal-to-noise ratio of $(\sigma_\beta^2\sigma_X^2 + \sigma_{\delta y^*}^2)/\sigma_\varepsilon^2 = 10$, and a regularization parameter of $\lambda=10^{-6}$.
We use a linear teacher model $y^*(\vbx) = \vbx\cdot\vbbeta$ ($\sigma_{\delta y^*}^2 = 0$) for all plots.

To find the solution for a particular regression problem, we solve a different (but equivalent) system of equations depending on whether $N_p < M$ or $N_p > M$, allowing us to reduce the size of the linear system we need to solve.
If $N_p < M$, we solve the system of $N_p$ equations
\begin{align}
\qty[\lambda I_{N_p} + Z^TZ]\hbw &=  Z^T\vby
\end{align}
for the  $N_p$ unknown fit parameters $\hbw$ where $I_{N_p}$ is the $N_p\times N_p$ identity matrix. 
This equation is identical to that in Eq.~\eqref{eq:exact} in the main text.

Alternatively, if $N_p > M$ we solve a system of $M$ equations,
\begin{align}
\qty[\lambda I_M + ZZ^T]\hba &= \vby,
\end{align}
for the $M$ unknowns $\hba$ where $I_M$ is the $M\times M$ identity matrix.
We then convert to fit parameters via the formula $\hbw = Z^T \hba$.

\subsection{Bias-Variance Decompositions}

To efficiently calculate the ensemble-averaged bias and variance, we take inspiration from Eq.~\eqref{eq:biastwodatasets}.
During each simulation, we independently generate two training data sets $\mathcal{D}_1$ and $\mathcal{D}_2$.
Using the results from the first training set, we calculate the training and test error.
To calculate the bias, we also calculate the label predictions for both training sets for an identical test set, $\hby_1$ and $\hby_2$, and record the residual label errors between these predictions and the true labels of the test set $\vby^{*\prime}$ We then record the dot product $(\hby_1-\vby^{*\prime})\cdot(\hby_2-\vby^{*\prime})$. When averaged over many simulations, this quantity approximates the bias. We can then subtract this quantity from the average test error to find the variance.

\subsection{Eigenvalue Decompositions of Kernel Matrices}

For each of the numerical eigenvalue distributions for the kernel matrices presented in the main text, we choose $M = 4096$.
We then average over the distributions for 10 independently sampled matrices when  $\alpha_p = 1$ or $\alpha_p = 8$ and over 80 matrices when  $\alpha_p = 1/8$.
In this way, we ensure that the same number of non-zero eigenvalues is present in the part of the histograms corresponding to the bulk of the distributions (the distribution excluding the delta function at zero).
For $M < N_p$ we calculate the eigenvalues of $Z^TZ$, while for $M > N_p$ we instead calculate the eigenvalues of $ZZ^T$ since this matrix is smaller and contains the same non-zero eigenvalues. In the later case, we then manally append an additional $N_p -M$ zero-valued eigenvalues to the distribution.

\end{document}

%% file: manuscript_rlfm_bias_var.bbl
%